\documentclass[lettersize,journal]{IEEEtran}
\usepackage{amsmath,amsfonts}
\usepackage{algorithmic}
\usepackage{algorithm}
\usepackage{array}
\usepackage{textcomp}
\usepackage{stfloats}
\usepackage{url}
\usepackage{verbatim}
\usepackage{graphicx}
\usepackage{cite}
\usepackage{booktabs}
\usepackage{float}
\usepackage{makecell}
\usepackage{tabularx}
\usepackage{longtable}
\usepackage{multirow}
\usepackage{threeparttable}

\ifCLASSOPTIONcompsoc
  \usepackage[caption=false,font=normalsize,labelfont=sf,textfont=sf]{subfig}
\else
  \usepackage[caption=false,font=footnotesize]{subfig}
\fi

\begin{document}

\title{LAMBDA: Covering the Multimodal Critical Scenarios for Automated Driving Systems by Search Space Quantization}

\author{Xinzheng Wu, Junyi Chen, Xingyu Xing, Jian Sun, Ye Tian,~\IEEEmembership{Member,~IEEE,} Lihao Liu and Yong Shen
\thanks{This work was supported by the National Key Research and Development Program of China under Grant 2022YFB2503001, the National Natural Science Foundation of China under Grant 52232015 and the Fundamental Research Funds for the Central Universities. (\textit{Corresponding author: Junyi Chen})}
\thanks{Xinzheng Wu, Junyi Chen, Xingyu Xing, Lihao Liu and Yong Shen are with the School of Automitive Studies, Tongji University, Shanghai 201804, China (e-mail: chenjunyi@tongji.edu.cn).}
\thanks{Jian Sun and Ye Tian are with the Key Laboratory of Road and Traffic Engineering, Ministry of Education, Department of Traffic Engineering, Tongji University, Shanghai 201804, China.}
}


\maketitle

\begin{abstract}
Scenario-based virtual testing is one of the most significant methods to test and evaluate the safety of automated driving systems (ADSs). However, it is impractical to enumerate all concrete scenarios in a logical scenario space and test them exhaustively. Recently, Black-Box Optimization (BBO) was introduced to accelerate the scenario-based test of ADSs by utilizing the historical test information to generate new test cases. However, a single optimum found by the BBO algorithm is insufficient for the purpose of a comprehensive safety evaluation of ADSs in a logical scenario. In fact, all the subspaces representing danger in the logical scenario space, rather than only the most critical concrete scenario, play a more significant role for the safety evaluation. Covering as many of the critical concrete scenarios in a logical scenario space through a limited number of tests is defined as the Black-Box Coverage (BBC) problem in this paper. We formalized this problem in a sample-based search paradigm and constructed a coverage criterion with Confusion Matrix Analysis. Furthermore, we propose LAMBDA (Latent-Action Monte-Carlo Beam Search with Density Adaption) to solve BBC problems. LAMBDA can quickly focus on critical subspaces by recursively partitioning the logical scenario space into accepted and rejected parts. Compared with its predecessor LaMCTS, LAMBDA introduces sampling density to overcome the sampling bias from optimization and Beam Search to obtain more parallelizability. Experimental results show that LAMBDA achieves state-of-the-art performance among all baselines and can reach at most 33 and 6000 times faster than Random Search to get 95\% coverage of the critical areas in 2- and 5-dimensional synthetic functions, respectively. Experiments also demonstrate that LAMBDA has a promising future in the safety evaluation of ADSs in virtual tests.
\end{abstract}


\section{Introduction}
\IEEEPARstart{T}HE past few years have witnessed the rapid development of highly automated driving technology, which has shown great potential in terms of accidents reduction, energy savings and congestion mitigation \cite{vahidi2018energy, xie2022heterogeneous}. However, the safety of automated driving systems (ADSs) remains a concern hindering the large-scale deployment of ADSs, which should be ensured by thorough tests \cite{iso26262, iso21448}. 

Recently, scenario-based virtual testing has become more and more significant in safety evaluation of ADSs due to its high efficiency, high flexibility, and low cost \cite{sun2022scenario}. Based on the three levels of scenario abstraction proposed by Menzel \MakeLowercase{\textit{et al.}} \cite{menzel2018scenarios}, a concrete test scenario can be obtained by choosing a specific value for each parameter of the logical scenario (namely sampling a point in the logical scenario space). Considering a logical scenario controlled by $m$ variables, each variable has $n$ possible values. Then the total number of concrete scenarios in this scenario space is $n^{m}$. It is obvious that testing these concrete scenarios exhaustively becomes impractical quickly with the growth of $m$ and $n$ (known as the "curse of dimensionality"). On the other hand, only a limited number of scenarios out of $n^{m}$ that are critical enough to trap the ADSs into danger deserve to be tested (known as the "long-tailed recognition").

One direct way to address these issues is to randomly sample in the logical scenario space to reduce the number of test scenarios. Among them, Monte Carlo sampling is the most classical method. Additionally, Latin Hypercube sampling (LHS) \cite{batsch2019performance} divides the search space into several even grids before sampling, ensuring that the search space is uniformly covered. Furthermore, Markov Chain Monte Carlo sampling \cite{akagi2019risk} takes into account the parameter distribution and covariance to make the sampling more efficient. However, the above methods can be inefficient with a low probability of critical scenarios under high-dimensional conditions.

Another category of methods introduces optimization algorithms to improve testing efficiency. Instead of generating all test scenarios at once, optimization algorithms generate only a limited number of scenarios at first and then conduct the testing. Based on the test results, an initial understanding of the distribution of critical scenarios can be obtained which will guide the generation of the next batch of scenarios to the most likely critical areas. By doing this iteratively, the attention of sampling can quickly and adaptively converge on critical scenarios, forming a method called adaptive sampling \cite{sun2022adaptive}. During the above process, one test procedure for an ADS is abstracted into a black-box objective function $f$, whose input is a vector $\boldsymbol{x}$ ($\boldsymbol{x} \in \Omega$ ,$\Omega$ is the logical scenario space) representing the concrete scenario parameters, and the output is a scalar $y$ representing the scenario risk (aka scenario criticality). This is a typical Black-Box Optimization (BBO) problem that can be denoted as Eq. (\ref{eq1}). A comprehensive review of the BBO algorithms and their applications in safety evaluation of ADSs are summarized in Section \uppercase\expandafter{\romannumeral2}.

\begin{equation}\label{eq1}
   \boldsymbol{x}^* = \mathop{\arg\max}\limits_{\boldsymbol{x}}f(\boldsymbol{x}), \boldsymbol{x} \in \Omega
\end{equation}

\begin{figure}[t] 
      \centering
      \includegraphics[width=8cm]{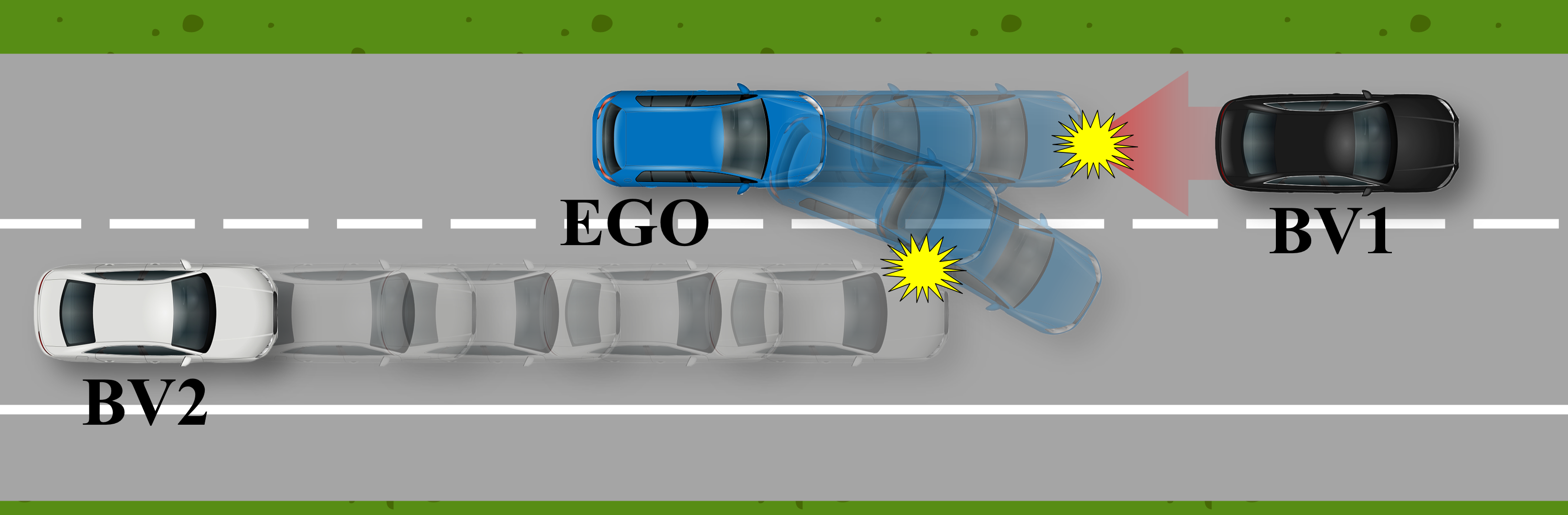}
      \caption{An example to show the multimodal nature of the distribution of critical scenarios.}
      \label{multimodalRisk}
\end{figure}

Following the idea of finding scenarios as critical as possible, BBO algorithms are able to falsify the safety of an ADS (that is, to prove that the system is unsafe when counterexamples that violate safety requirements are found) \cite{riedmaier2020survey}. However, due to the high complexity and dimensionality of real-world scenarios, the distribution of critical concrete scenarios is usually multimodal \cite{ding2021multimodal}. An example is shown in Fig. \ref{multimodalRisk}, from the perspective of EGO, accidents can occur when the front BV1 suddenly brakes or when EGO performs an evasive maneuver and collides with BV2 behind it. The two events are controlled by a series of scenario parameters that are distributed separately in the entire logical scenario space. As a result, it is clear that only seeking the most critical scenarios can neither guarantee the safety of the ADS nor provide a comprehensive and overall evaluation of the ADSs' safety performance under a logical scenario.

To fill the gap, the key is to ensure coverage of the critical scenarios of the BBO algorithms. Given a tolerable scenario risk criterion $\delta$, all critical scenarios can be defined with an inequality as $f(\boldsymbol{x}) > \delta, \boldsymbol{x} \in \Omega $, which is called the black-box safety inequality in this paper. With the inequality in mind, the Black-Box Coverage (BBC) problem emerges: how to identify all the solution sets of a black-box inequality with limited evaluation opportunities. In the current literature, although numerous mechanisms have been developed to avoid BBO algorithms being trapped into local optima, it remains an open question to avoid BBO algorithms being trapped into one particular global optimum when facing a multimodal objective function, where any scenario that satisfies the black-box safety inequality can be considered globally optimal. 
In this paper, we developed an algorithm named LAMBDA (\textbf{L}atent-\textbf{A}ction \textbf{M}onte-Carlo \textbf{B}eam Search with \textbf{D}ensity \textbf{A}daption) to solve BBC problems. LAMBDA can quickly focus on the possible solution set by recursively partitioning the search space into good and bad subspaces and always give a glimpse into the under-explored but potentially promising area with a modified UCB (Upper Confidence Bound) bandit algorithm to guarantee the coverage of the critical scenarios. 

The main contributions of this paper can be summarized as follows: 
\begin{enumerate}
\item{The formalization to transfer the sample-based safety evaluation of ADSs in logical scenarios into the BBC problem, as well as a metric to evaluate the coverage of the logical scenario based on the Confusion Matrix Analysis.}
\item{The LAMBDA algorithm for solving BBC problems with an adaptive kernel density estimator (KDE) for UCB score calculation to overcome the sampling bias, and its adaptation methods to other local samplers.}
\item{A high-dimensional multimodal synthetic function that specifically serves as a black-box objective function for the BBC problem, and benchmarks on 2- and 5-dimensional synthetic functions to verify the effectiveness of the proposed algorithm compared with other baseline algorithms.}
\item{A practical issue for safety evaluation of ADSs to show the application of our algorithm based on Software-in-the-Loop (SiL) simulation testing.}
\end{enumerate}

The reminder of this paper is organized as follows: Section \uppercase\expandafter{\romannumeral2} reviews the current studies of the BBO algorithms and the optimization-based safety evaluation methods of ADS. The definition of the BBC problem and a metric to evaluate the performance of algorithms under the BBC problem are proposed in Section \uppercase\expandafter{\romannumeral3}. In Section \uppercase\expandafter{\romannumeral4}, the framework, the improvements and the adaptation methods of LAMBDA are introduced. Three experiments are then conducted in Section \uppercase\expandafter{\romannumeral5} to verify our proposed algorithms in low-dimensional, high-dimensional, and practical cases. Finally, this paper is summarized in Section \uppercase\expandafter{\romannumeral6}.

\section{Related Works}
In this section, we first review the BBO algorithms proposed in the current literature and choose several algorithms as baselines for the benchmarks. The studies on the safety evaluation of ADSs based on optimization are then summarized and research gaps are discussed.

\subsection{The black-box Optimization Methods}

Current BBO algorithms can be divided into 3 categories: population-based, surrogate-based, and Monte Carlo Tree Search (MCTS)-based. Population-based methods are mainly inspired by the behavior of the biological population. Genetic Algorithm (GA) \cite{holland1992adaptation} simulates evolution and selection by using cross and mutation operators to propose new samples. Differential Evolution (DE) \cite{storn1997differential} is similar to GA but uses vector differences to perturbate current samples. Besides, Particle Swarm Optimization (PSO) \cite{kennedy1995particle}, CMA-ES \cite{hansen2003reducing}, Simulated Annealing \cite{kirkpatrick1983optimization} are also widely known in population-based optimization methods. In this paper, we choose the most classic GA and DE as baselines in benchmarks from the population-based methods.

Surrogate-based methods maintain a surrogate model of the objective function during optimization and determine candidates to be evaluated obeying to the surrogate. Bayesian Optimization (BO) \cite{pelikan1999boa} is a typical surrogate-based method widely used in many fields such as Neural Architecture Search (NAS) \cite{white2021bananas}, structure design \cite{yamawaki2018multifunctional}, and hyper-parameter tunning \cite{snoek2012practical}. The most significant limitation of BO is the $O(n^3)$ complexity of the Gaussian Process Regressor (GPR) surrogate, leading to inefficacy in problems of more than 10 dimensions and several thousand evaluations \cite{dalibard2017framework}. To address this issue, Tree-structured Parzen Estimator (TPE) \cite{bergstra2011algorithms} and BOHB \cite{falkner2018bohb} replace the GPR surrogate with Parzen Estimator to obtain higher performance. Meanwhile, TuRBO \cite{eriksson2019scalable} introduces trust region, implicit multi-armed bandit and a restart mechanism to adapt BO to large scale. In this paper, we compare LAMBDA with BO and TuRBO.

MCTS-based methods are most relevant to our work. Despite its fantastic performance in Go \cite{silver2016mastering}, MCTS can be used to solve optimization problems. The search space partition is a common technique for adapting MCTS to continuous problems in this line of research. Deterministic Optimistic Optimization (DOO), Simultaneous Optimistic Optimization (SOO) \cite{munos2011optimistic}, and Hierarchical Optimistic Optimization (HOO) \cite{Bubeck2011XArmedB} use k-ary partitions, while Voronoi Optimistic Optimization (VOO) \cite{kim2020monte} introduces more efficient Voronoi partitions. Recently, Wang \MakeLowercase{\textit{et al.}} proposed LaNAS \cite{wang2021sample} and LaMCTS \cite{wang2020learning} to learn the arbitrary decision boundaries of partitions. Moreover, Wang \MakeLowercase{\textit{et al.}} \cite{wang2021sample} showed that linear partitions perform equally well or better than curving partitions. And Yee \MakeLowercase{\textit{et al.}} \cite{yee2016monte} proposed the KR-UCT which adapts MCTS to the continuous action space with execution uncertainty using the kernel density. The proposed LAMBDA in this paper inherits the above thoughts of search space partition and further introduces sampling density information to overcome the sampling bias of optimization. For MCTS-based methods, we take the latest and most state-of-the-art LaMCTS into benchmarks.

\subsection{Optimization-based Safety Evaluation Methods of ADS}

Safety evaluation of ADSs has been a popular research topic for years, mainly conducted through scenario-based tests, in the form of finding or generating critical scenarios \cite{ding2023survey}. Critical scenarios can be generated through data-driven \cite{Ding2023CausalAFCA}, adversary-based \cite{chen2022adversarial}, and sample-based methods, while this paper focuses on sample-based methods, especially optimization-based sampling methods. 

Currently, numerous studies are introducing the BBO algorithms mentioned above into scenario-based testing of ADSs to find the most critical scenario and conduct the pressure test. For population-based methods, Klischat \MakeLowercase{\textit{et al.}} \cite{klischat2019generating} applied DE and PSO to generate critical scenarios in complex situations (for example, intersections). 
Kl{\"u}ck \MakeLowercase{\textit{et al.}} \cite{kluck2019performance} compared the performance of GA with simulated annealing and random testing for optimization of test parameters. Li \MakeLowercase{\textit{et al.}} \cite{li2020av} combined GA with a local fuzzer to perturb the parameters of the trajectories of other traffic participants to create safety-hazardous situations. When it comes to surrogate-based methods, Gladisch \MakeLowercase{\textit{et al.}} \cite{gladisch2019experience} used Systems Theoretic Process Analysis (STPA) to initiate a test function scenario and then applied BO to find critical combinations of test parameters and their values. Sun \MakeLowercase{\textit{et al.}} \cite{sun2022adaptive} compared six surrogate models with two logical scenarios and found that Extreme Gradient Boosting (XGB) stood out among them. Based on an information-theoretic approach, Gong \MakeLowercase{\textit{et al.}} \cite{gong2023adaptive} proposed a novel acquisition function for a GPR-based multi-fidelity sampling framework to reduce the cost of CAV safety evaluation. As for the MCTS-based methods, Koren \MakeLowercase{\textit{et al.}} \cite{koren2018adaptive} used MCTS and Deep Reinforcement Learning (DRL) to find collision scenarios, respectively, and then compare their efficiency.

As discussed in Section \uppercase\expandafter{\romannumeral1}, the studies reviewed above are essentially falsification methods, which means that they focus only on one or several optimal solutions and the efficiency of the applied algorithm, leading to a biased search result that cannot validate the comprehensive safety performance of ADSs. To fill the gap, researchers have taken into account test coverage, with the goal of balancing exploration and exploitation. Feng \MakeLowercase{\textit{et al.}} \cite{feng2022multimodal} modified PSO to enhance the exploration ability and adopted the index F1 to quantify the coverage of critical scenarios. Wang \MakeLowercase{\textit{et al.}} \cite{wang2022comprehensive} proposed a modified expected improvement criterion that emphasized more on the coverage of failure mode as an acquisition function for a GPR-based adaptive sampling, while utilizing contextual information to encourage exploration. From another perspective, Zhu \MakeLowercase{\textit{et al.}} \cite{zhu2022hazardous} attempted to identify a safety performance boundary (SPB) with the sampled data using the Gaussian distribution method to cover all hazardous parameter spaces. Following the same idea, Wang \MakeLowercase{\textit{et al.}} \cite{wang2022safety} utilized an optimized gradient descent searching algorithm to quickly and accurately identify SPB and further proposed the concept of safety tolerance performance boundary (STPB).
However, it is still a concern to guarantee both the efficiency and coverage of these algorithms under multimodal and high-dimensional BBC problems. This paper proposes a metric for evaluation, an algorithm for solution, as well as a synthetic function for verification to address this issue.

\section{Definition and Metric of the BBC Problem}
\subsection{Definition}

The BBC problem can be defined as follows. Given a black-box function $f$ which represents method to obtain the criticality of a scenario and a criterion $\delta$, the goal is to figure out the solution set $\boldsymbol{x}$ satisfying the inequality $f(\boldsymbol{x})>\delta$ and achieve as much coverage of $\boldsymbol{x}$ as possible. In this paper, only deterministic functions are considered. The key question of BBC problem is that no information else except the input and output of $f$ can be obtained, and the number of times to evaluate $f$ is limited. It is a typical search-based optimization problem that might be solved within an optimization paradigm. Table \ref{tab1} shows the denotations in this paper.

To be specific, in a BBC problem, the optimization agent $\Pi$ has a limited optimization budget $N$, which means that the agent only has $N$ opportunities to sample from $f$. At each sampling from $f$, the agent decides where to place the sampling point $\boldsymbol{x}$ according to the historical sample records $\mathcal{D}$, then gets an output $y=f(\boldsymbol{x})$ through evaluating $f$, and appends the new pair $(\boldsymbol{x},y)$ to $D$ , as shown in Eq. (\ref{eq3}). After running out the optimization budget, agent $\Pi$ gets sample records $\mathcal{D}$ of size $N$.

\begin{equation}\label{eq3}
\begin{aligned}
   (\boldsymbol{x},y)&:=(\Pi(\mathcal{D}),f(\boldsymbol{x}))\\
   \mathcal{D}&:=\mathcal{D}\cup(\boldsymbol{x},y)
\end{aligned}
\end{equation}

\begin{table*}[t]
\renewcommand{\arraystretch}{1.3}
\caption{Denotations\label{tab1}}
\centering
\begin{tabular}{ p{0.1\linewidth}  p{0.35\linewidth} p{0.1\linewidth} p{0.35\linewidth}}
\toprule
\textbf{Notation} & \textbf{Definition} & \textbf{Notation} & \textbf{Definition}\\
\midrule
$\Omega$ & whole search space & $\Omega_i$ & subspace of node $i$\\

$f$ & objective function & $\hat{f}$ & regressor of the objective function\\

$(\boldsymbol{x},y)$ & the pair of a sample point and its result & $\delta$ & criterion in the inequality $f(\boldsymbol{x})>\delta$\\

$\Omega_\delta$ & subspace satisfying $f(\boldsymbol{x})>\delta$ & $\Omega_{\lnot\delta}$ & subspace not satisfying $f(\boldsymbol{x})>\delta$\\

$\widehat{\Omega}_\delta$ & predicted subspace satisfying $f(\boldsymbol{x})>\delta$ & $\widehat{\Omega}_{\lnot\delta}$ & predicted subspace not satisfying $f(\boldsymbol{x})>\delta$\\

$\mathcal{D}$ & set of historical sample records & $\mathcal{D}_i$ & set of historical sample records located in $\Omega_i$\\

$X$ & set of historical sample points & $X_i$ & set of historical sample points located in $\Omega_i$\\

$Y$ & set of historical sample results & $Y_i$ & set of historical sample results located in $\Omega_i$\\

$N$ & optimization budget & $n_i$ & number of samples in $\Omega_i$\\

$v_i$ & average value of $\Omega_i$ & $\rho_i$ & average sampling density of $\Omega_i$\\

$c_p$ & exploration factor $\Omega_i$ & $w_i(\boldsymbol{x})$ & normalized weight of sample point $x$ in $\Omega_i$\\

$\widetilde{\boldsymbol{x}}$ & set of validation points & & \\
\bottomrule
\end{tabular}
\end{table*}

\subsection{Metric} \label{metric}
\begin{figure}[!b] 
      \centering
      \includegraphics[width=\linewidth]{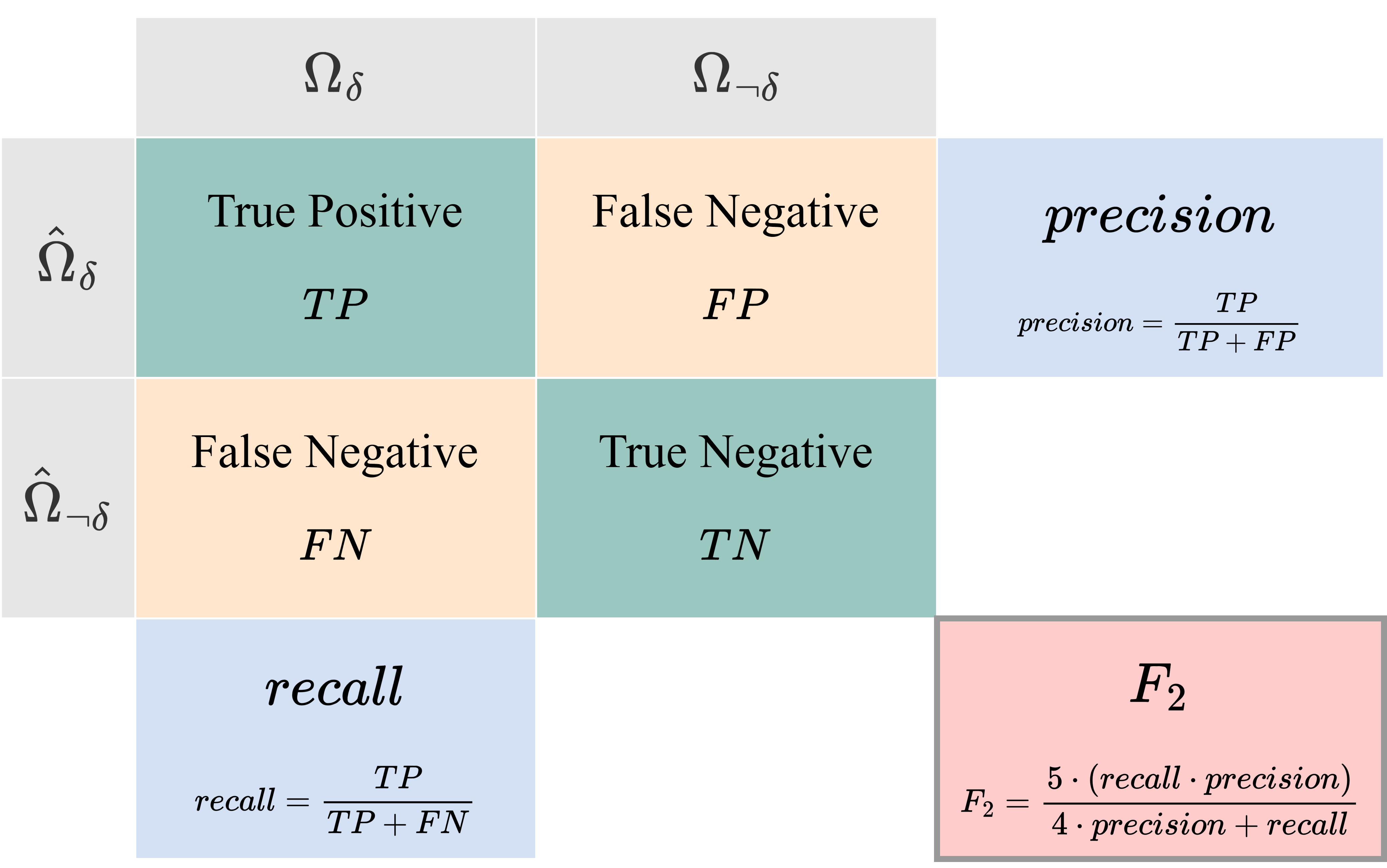}
      \caption{The confusion matrix of BBC problem.}
      \label{f2_CondusionMatrix}
\end{figure}

As mentioned above, different from that in a BBO problem which is to evaluate whether the optimal value is found, the goal of the metric in a BBC problem is to evaluate how many critical values are covered. For this purpose, we use a classifier $\mathcal{C}$ trained from sample records $\mathcal{D}$ to predict whether $\widetilde{\boldsymbol{x}}$ meets $f(\widetilde{\boldsymbol{x}})>\delta$, and further divide the search space $\Omega$ into subspace $\widehat{\Omega}_\delta$ and subspace $\widehat{\Omega}_{\lnot\delta}$. Note that $\widetilde{\boldsymbol{x}}$ is the set of validation points generated through gird sampling in $\Omega$, and $\widehat{\Omega}_\delta$ is the subspace predicted by $\mathcal{C}$ that satisfies $f(\widetilde{\boldsymbol{x}})>\delta$, namely the set of critical scenarios, while $\widehat{\Omega}_{\lnot\delta}$ is $\Omega-\widehat{\Omega}_\delta$. With the actual $\Omega_\delta$ and $\Omega_{\lnot\delta}$ known, the performance of $\mathcal{C}$, which is directly related to the quality of $\mathcal{D}$, can be assessed with Confusion Matrix Analysis \cite{powers2020evaluation}. Since $\mathcal{D}$ is the sample records of optimization agent $\Pi$, the ability to cover all the critical scenarios of $\Pi$ can be reflected this way. The confusion matrix of the BBC problem is illustrated in Fig. \ref{f2_CondusionMatrix}.

As shown in Fig. \ref{f2_CondusionMatrix}, in the Confusion Matrix, $recall$ represents the coverage of $\mathcal{C}$’s prediction to the solution set of $f(\widetilde{\boldsymbol{x}})>\delta$. However, an intentional classifier can decide that all points in the search space must be satisfying $f(\widetilde{\boldsymbol{x}})>\delta$, then the recall would be 100\%. For all practical purposes, it does not make sense. Thus, the $precision$ must be taken into account. In this paper, the $F_2$ score is used to balance $recall$ and $precision$, with $recall$ being more emphasized. Fig. \ref{f2_CondusionMatrix} illustrates the calculation methods for $recall$, $precision$, and $F_2$ as well as their relationships.

\begin{figure}[!b] 
      \centering
      \includegraphics[width=\linewidth]{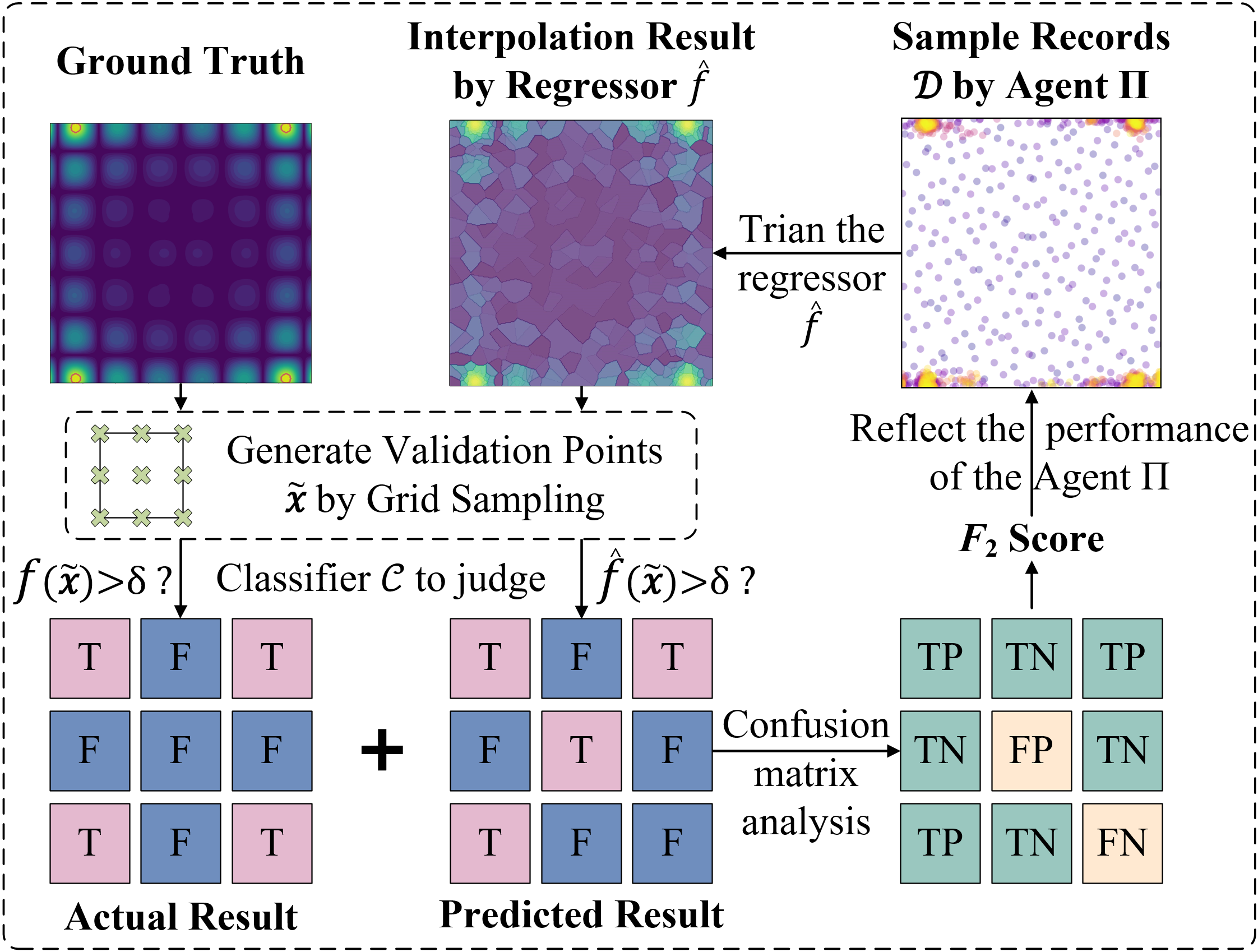}
      \caption{The evaluation process of BBC problem.}
      \label{F2_calculation}
\end{figure}

In practice, the key point of the BBC problem is how to acquire information about the search space with a limited optimization budget. Therefore, this paper mainly focuses on the optimization agent $\Pi$, which decides where to sample the search space and acquire information about $f$. As for the classifier $\mathcal{C}$, appropriate methods can be chosen according to the domain knowledge of each practical issue. In this paper, for generality and fairness, a basic configuration of $\mathcal{C}$ was used in all experiments. A regressor $\hat{f}$ of $f$ is built from $\mathcal{D}$ with the linear interpolation method of $SciPy$ \cite{2020SciPy-NMeth}. Thereby the classifier can be constructed with $\hat{f}$ simply, as shown in Eq.(\ref{eq4}). An illustration of the evaluation process can be seen in Fig. \ref{F2_calculation}.

\begin{equation}\label{eq4}
   \mathcal{C}(\widetilde{\boldsymbol{x}})=True \ \text{IF}\ \hat{f}(\widetilde{\boldsymbol{x}})>\delta\ \text{ELSE}\ False
\end{equation}

\section{LAMBDA Algorithm} \label{lambda}
In this section, we first give a brief introduction to LaMCTS, the predecessor of our proposed algorithm. Then the limitation of LaMCTS is demonstrated and our solutions are provided. Finally, the framework and details of our LAMBDA algorithm are depicted.

\subsection{Brief Introduction of LaMCTS}
MCTS is an algorithm to solve discrete sequential decision problems, widely known for its surprising performance in Go \cite{silver2016mastering}. However, MCTS is not originally applicable to continuous optimization because it requires a fixed and limited action space, which is usually arbitrary and infinite in continuous optimization problems. LaMCTS \cite{wang2020learning} breaks this limitation by learning partitions of the search space to form latent actions and adapts MCTS to optimization problems in continuous search space.

\begin{figure}[b] 
      \centering
      \includegraphics[width=\linewidth]{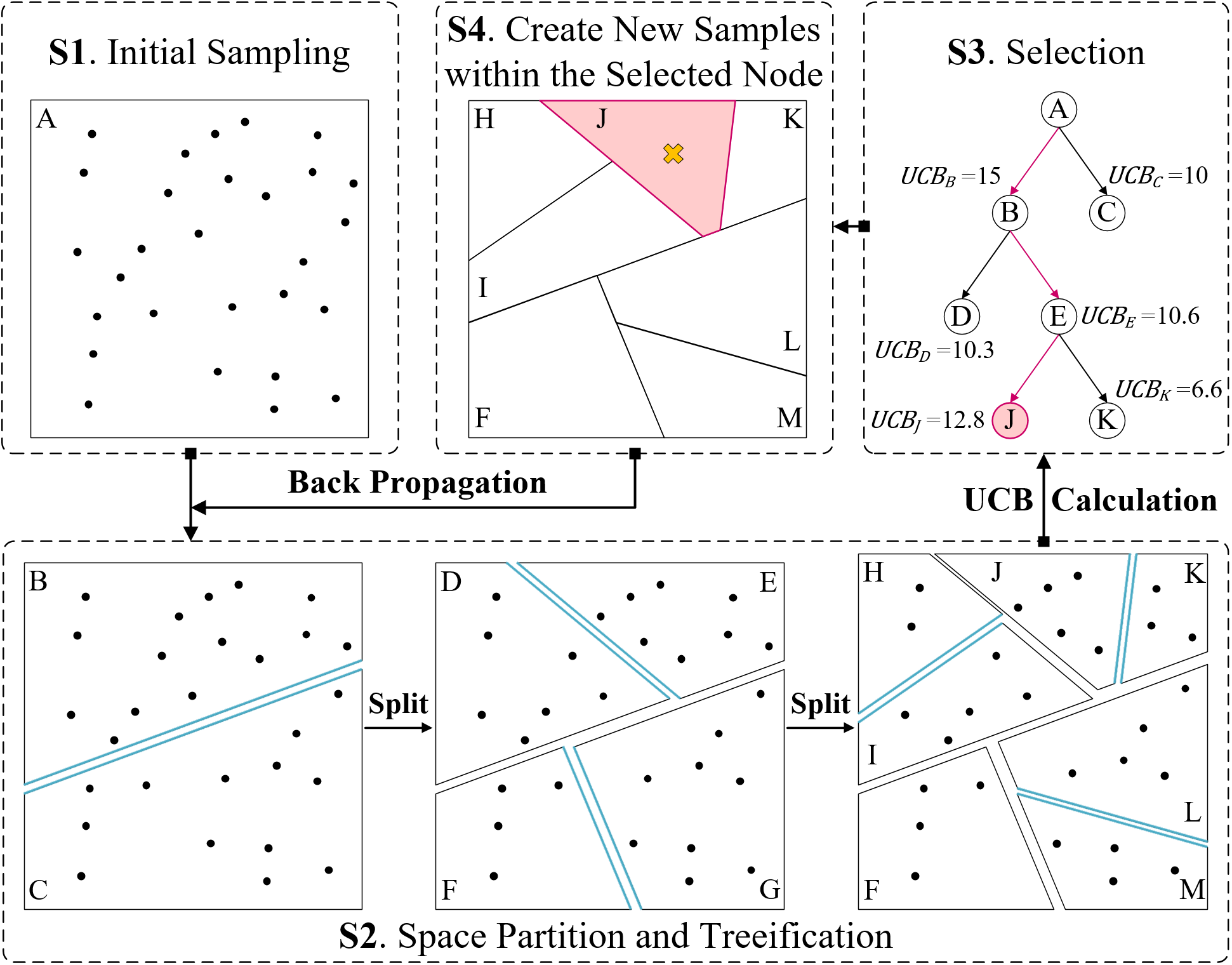}
      \caption{The workflow of LaMCTS.}
      \label{lamcts_workflow}
\end{figure}

The main steps of LaMCTS are summarized in Fig. \ref{lamcts_workflow}, where A represents the entire search space (i.e., the parent node), and other different letters represent the child nodes obtained as the partition proceeds. As depicted in Fig. \ref{lamcts_workflow}, after initial sampling from the entire search space, LaMCTS builds a tree by recursively splitting the search space into regions with high/low function values. Then, the UCB of each leaf node is calculated to guide the selection of the most promising node, within which new samples will be created. Finally, the function values of the newly created samples are obtained and back-propagated to the historical data set before the next iteration.

Among the above steps, the calculation of the UCB, which directly determines where to sample next, is of great importance. In \cite{wang2020learning}, the formula for calculating UCB (denoted as $UCB_1$) from a parent node $A$ to a child node $B$ is as shown in Eq. (\ref{eq5}).

\begin{equation}\label{eq5}
    UCB_1(A\rightarrow B)=\frac{\sum_{\boldsymbol{x}\in\mathcal{D}_B}f(\boldsymbol{\boldsymbol{x}})}{n_B}+2\cdot c_p \cdot \sqrt{\frac{2ln\ n_A}{n_B}}
\end{equation}
where $n_A$ and $n_B$ represent the number of historical samples located in node $A$ and node $B$, respectively. And $c_p$ is a tunable hyper-parameter to balance the exploitation and exploration of the algorithm.

\subsection{Sampling Bias of LaMCTS and Solutions}
\begin{figure}[t] 
      \centering
      \includegraphics[width=\linewidth]{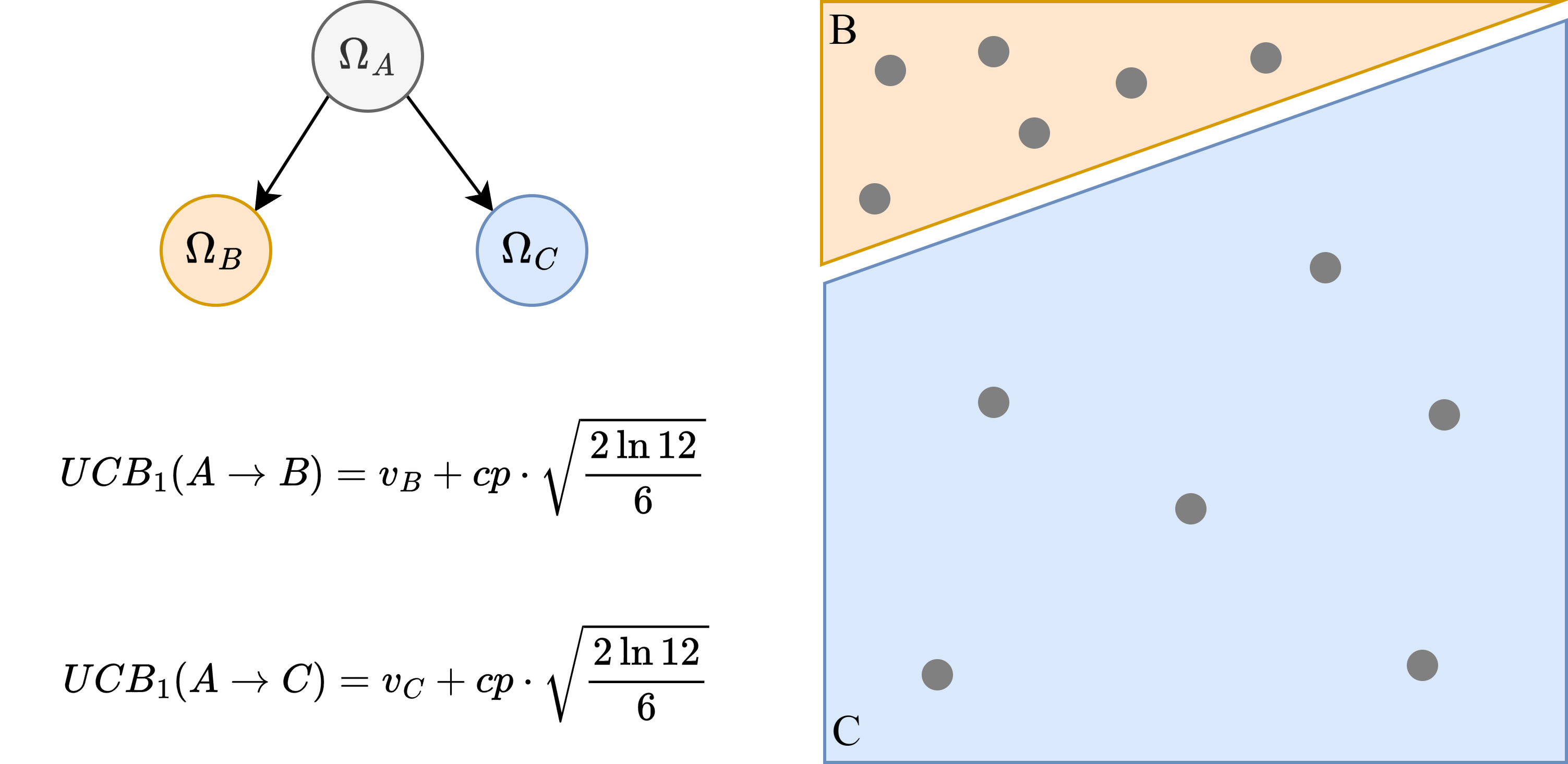}
      \caption{An intuitive example of sampling bias. }
      \label{sampling_bias}
\end{figure}

Despite the outstanding performance in the BBO problem, LaMCTS is faced with a further question called sampling bias when it comes to the BBC problem. As the optimization proceeds, samples concentrate towards more hopeful subspaces. Hence, the record set suffers from an imbalance of samples, leading to the following: A) the partition learned from $\mathcal{D}$ overemphasizes these hopeful regions and cuts them into many fragmentary subspaces. B) the UCB score based on the mean value and the number of samples in a subspace loses efficacy. The two impacts above make the optimization agent trapped in one local modality of the search space. Fig. \ref{sampling_bias} shows an intuitive example of this phenomenon.

As shown in Fig. \ref{sampling_bias}, assuming $v_B>v_C$, then $\Omega_B$ is the hopeful region of the search space. Compared to $\Omega_C$, with the same number of samples ($n_B=n_C=6$), $\Omega_B$ has a higher mean value of samples but a smaller volume. Based on Eq. (\ref{eq5}), the exploration terms constructed with the number of samples turn out to be equal and no longer encourage sampling in the under-explored subspace $\Omega_C$. To handle this issue, in this paper, density information is introduced to overcome the sampling bias. The Inverse Probability Weighting (IPW) technique is used to rebalance the sample records in search space partitioning and UCB calculation, which will be detailed later.

In order to estimate the sampling density efficiently, we developed an adaptive Kernel Density Estimator (KDE) with Approximate Nearest Neighborhood (ANN). The kernel bandwidth is dynamically calculated by querying the $k$-th nearest neighbors’ minimal bounding sphere so that the KDE can adapt to different problems without tuning hyper-parameters. In addition, $faiss$ \cite{johnson2019billion} was utilized to improve retrieval efficiency. The speedup structure (or Index in $faiss$) needs to be retrained periodically, because ANN assumes that data are drawn from one fixed distribution while the distribution of samples gradually concentrates to target subspaces during the optimization.

\subsection{Framework and Improvements of LAMBDA}

The framework of LAMBDA is similar to that of LaMCTS, consisting of extension (treeification), selection, simulation, back-propagation, etc. The structure of LAMBDA is shown in Fig. \ref{LambdaProcess}. Note that the F2 score calculation module for evaluation is an independent module decoupled from the entire search process. The calculation can be performed at any search stage of any optimization algorithm as long as the historical samples and the black-box objective function values of the validation points are available.

\begin{itemize}
\item{\textbf{Initial Sampling}. Sample an initial $\mathcal{D}$ to construct the partition tree. Sobol sequence \cite{sobol1967distribution} was used in sampling to get more uniformity.}

\begin{figure}[!b] 
      \centering
      \includegraphics[width=\linewidth]{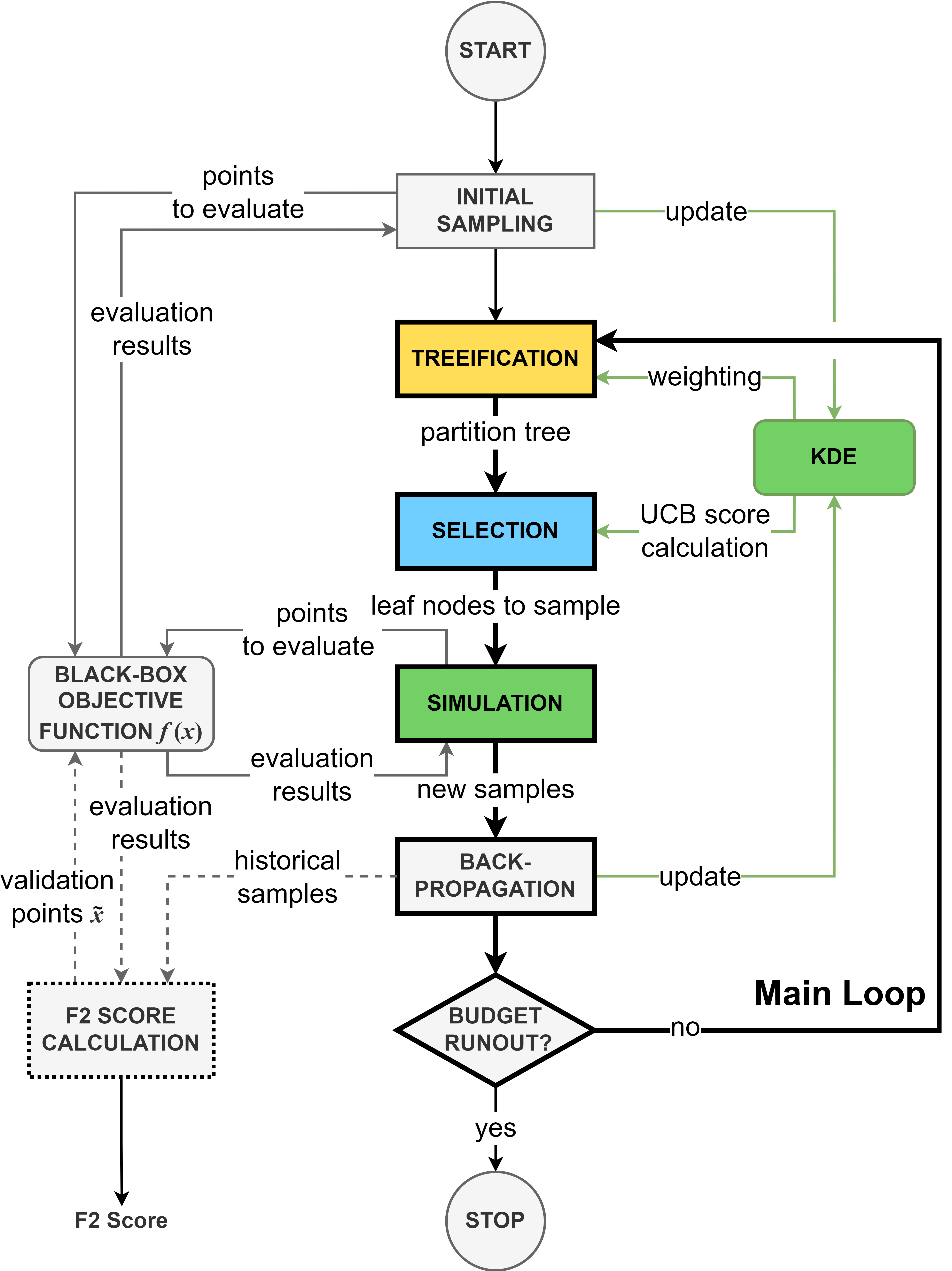}
      \caption{The structure of LAMBDA.}
      \label{LambdaProcess}
\end{figure}

\item{\textbf{Treeification}. Treeify the search space into subspaces recursively to form a quantization of the search space. Each leaf node in the partition tree represents a subspace. In this paper, we used linear partitions. More complex forms of partitions can be used, but with a greater calculation burden in the treeification and simulation stages. Sampling density estimated by KDE is introduced into this stage to get unbiased partitions.}
\item{\textbf{Selection}. Select a batch of subspaces to sample according to the UCB scores of the nodes in the partition tree. A new UCB calculation method is proposed based on the sampling density to overcome sampling bias.}
\item{\textbf{Simulation}. Sample from the selected subspaces. For BBC problems of less than 3 dimensions, LAMBDA just picks a random point in a subspace with rejected sampling (if a random point is judged inside the subspace by SVM, we accept it; otherwise, we reject it). The efficiency of rejected sampling can be accepted due to the linear partitions and low dimension. However, when it comes to high-dimensional problems, advanced local samplers should be applied, which will be detailed in Section \ref{sample_with_others}.}
\item{\textbf{Back-Propagation}. Back-propagate new samples to update the partition tree. The partition tree is reconstructed every period, thus back-propagation is needed within each period.}
\end{itemize}

The main improvements of LAMBDA are as follows.

\subsubsection{Latent Action}

A latent action is the partition of a search space or subspace learned from historical samples that forms a decision boundary to split the parent space into two children, the good and the bad. The good one holds samples with higher objective function values. By recursively splitting the search space, a tree structure of partition is built on the search space. Then LAMBDA could utilize the partition tree to decide which subspaces should be paid attention. Moreover, stop conditions are required to control the recursive partition process. We used $leafsize$ and $depth$ as stop conditions: A) if the samples in a subspace are less than $leafsize$, the partition in this subspace will stop. B) if the branch depth of a leaf node reaches $depth$, the partition on this node will stop.

\begin{figure}[b] 
      \centering
      \includegraphics[width=\linewidth]{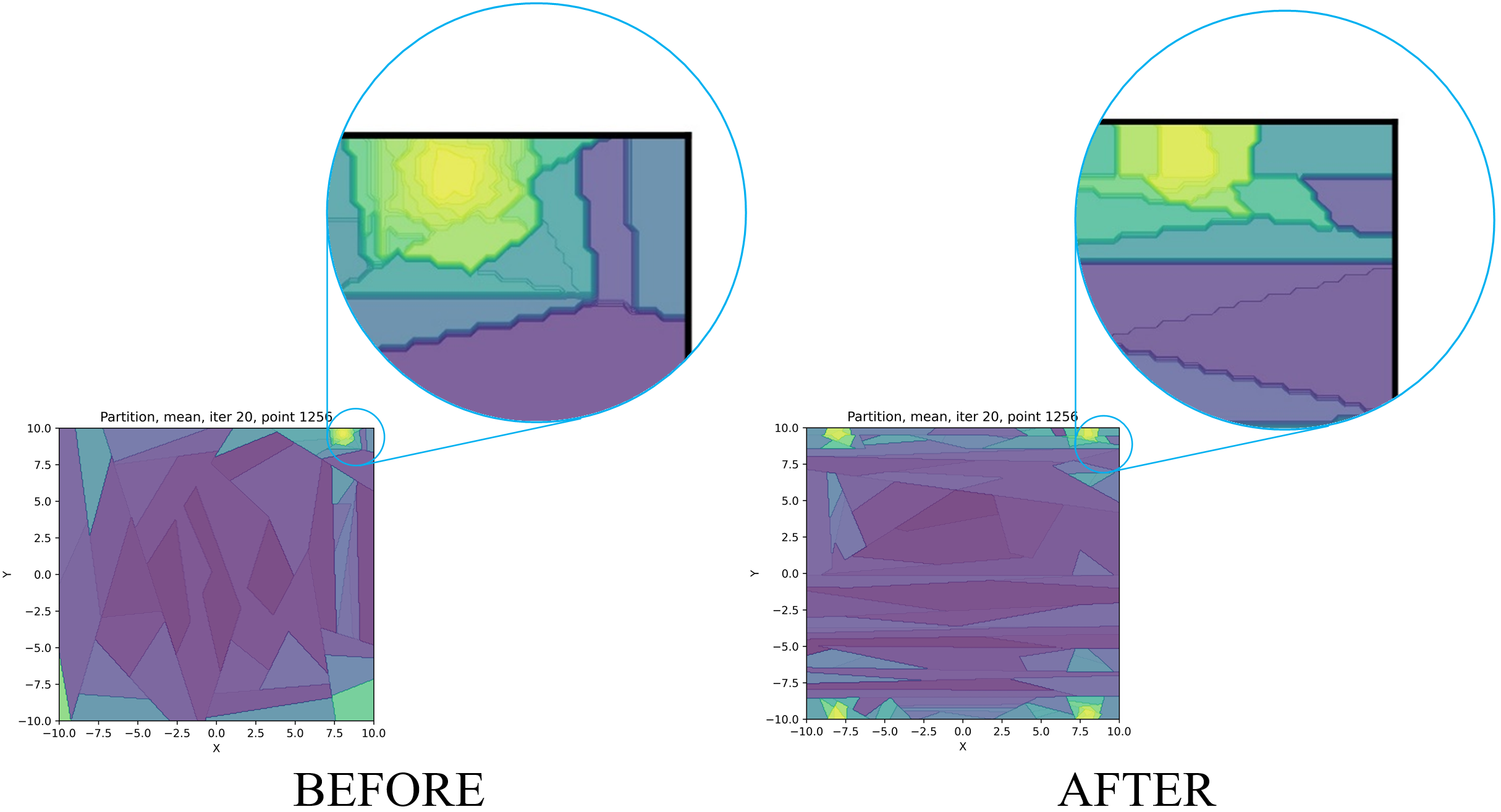}
      \caption{The comparison of partition between before and after weighting. The before one over-partitions the subspace at the top-right corner and forms many fragments.}
      \label{partation_Comparison}
\end{figure}

To split the search space, LAMBDA uses a KNN-SVM method to learn latent actions like LaMCTS \cite{wang2020learning}, which firstly clusters the samples into two groups to get pseudo labels, then uses pseudo labels to train an SVM classifier to get the decision boundary. However, LAMBDA weights the samples by the normalized inverse of each sample’s density to overcome the sampling bias. Samples in sparse areas get more importance than in dense areas. Fig. \ref{partation_Comparison} shows a comparison of partition between before and after weighting.

\subsubsection{Multi-Beam Search}
Beam Search is a technique broadly used in searching on graphs. Both the branch of best value and the other top-$k$ branches will be evaluated in Beam Search. It can improve the exploration of pure greedy search and parallelizability by $k$ times with the concurrent evaluation of $k$ branches.

\begin{figure}[t] 
      \centering
      \includegraphics[width=\linewidth]{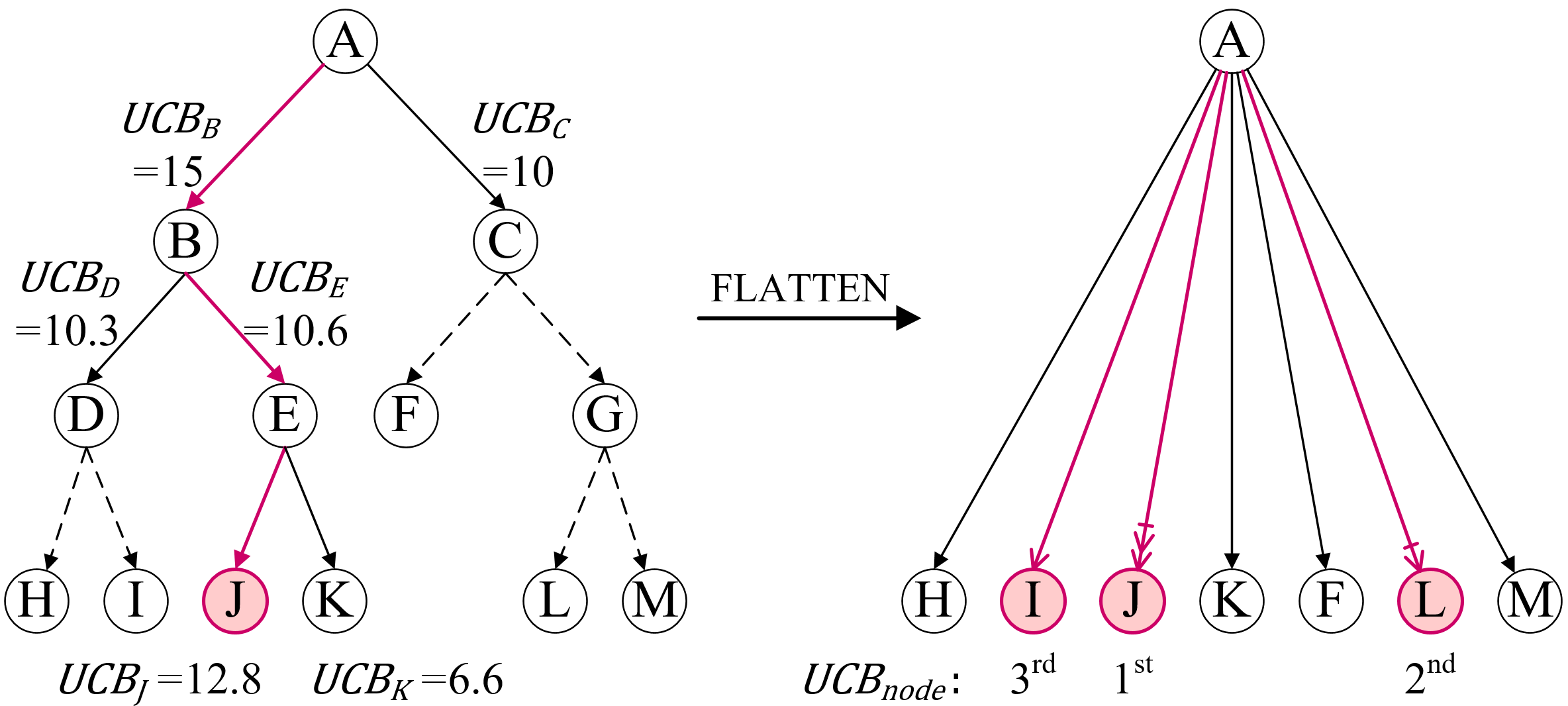}
      \caption{Flattening partition tree to perform beam search. Before flattening, only the subspace with the highest UCB score can be selected, while after flattening the top 3 branches (1st: J, 2nd: L, 3rd: I) can be selected and simulated simultaneously.}
      \label{comparisonBeamSearch}
\end{figure}

LAMBDA adapts the beam search to MCTS by flattening the partition tree, as shown in Fig. \ref{comparisonBeamSearch}. There are only root and leaf nodes in the flat partition tree. Therefore, when calculating UCB scores to perform selection, A) all parent nodes in the UCB expression should be the root node, B) top-$k$ branches of highest UCB scores should be selected all at once.

\subsubsection{Density Adaptive}
A new UCB calculation method was designed to replace the ineffective $UCB_1$(Eq. (\ref{eq5})) based on the mean value and the number of visits. The new UCB score is based on density, and we call it $UCB_\rho$. $UCB_\rho$ brings sampling density into account to rebalance the record set $\mathcal{D}$ and get an unbiased estimation of the exploitation term and the exploration term. Eq. (\ref{eq6}) describes the new UCB score.

\begin{equation}\label{eq6}
    UCB_\rho(A\rightarrow B)=\sum_{\boldsymbol{x} \in \mathcal{D}_B}f(\boldsymbol{x})w_B(\boldsymbol{x})+c_p \cdot log_{Adapt}\left(\frac{\overline{\rho}_A}{\overline{\rho}_B} \right)
\end{equation}
In which,
\begin{equation}\label{eq7}
w_B(\boldsymbol{x})=\frac{1/\rho(\boldsymbol{x})}{\sum_{\boldsymbol{x}\in \mathcal{D}_B} 1/\rho(\boldsymbol{x})} 
\end{equation}
\begin{equation}\label{eq8}
   \overline{\rho}_A = \sum _{\boldsymbol{x} \in \mathcal{D}_A} w_A(\boldsymbol{x})\rho(\boldsymbol{x}), \ \overline{\rho}_B = \sum _{\boldsymbol{x} \in \mathcal{D}_B} w_B(\boldsymbol{x})\rho(\boldsymbol{x})
\end{equation}
\begin{equation}\label{eq9}
   Adapt = {\frac{max(\overline{\rho}_{child})}{\overline{\rho}_A}}
\end{equation}
where $w_B(\boldsymbol{x})$ is the normalized weight at the historical sample point $\boldsymbol{x}$ in the subspace $\Omega_B$ considering the sampling density, and $\rho(\boldsymbol{x})$ represents the sampling density estimated by KDE. Eq. (\ref{eq7}) indicates that the higher the estimated density at $\boldsymbol{x}$, the lower the weight at that point. In Eq. (\ref{eq8}), $\overline{\rho}_A$ and $\overline{\rho}_B$ represent the mean sampling density of the parent subspace $\Omega_A$ and the child subspace $\Omega_B$, respectively. Note that here we estimate the mean sampling density of a subspace by utilizing in-hand samples from $\mathcal{D}$ with their weights calculated before, instead of sampling new points with Monte Carlo and then calculating their average value, to improve the computational efficiency. Furthermore, $Adapt$ is an adaptive base of the logarithm to limit its value range.

\begin{figure}[t] 
      \centering
      \includegraphics[width=\linewidth]{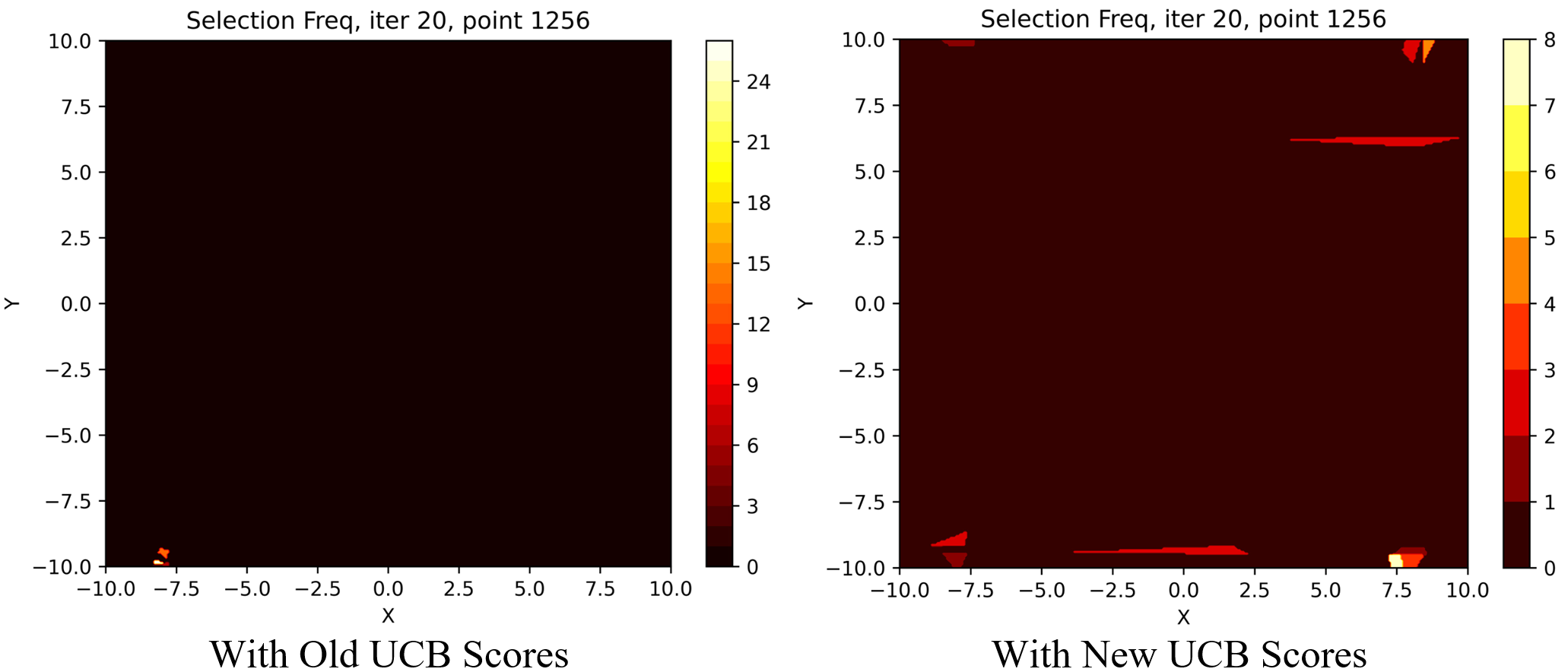}
      \caption{Sampling frequency with old and new UCB scores. Samples over-concentrate on one tiny point with the old UCB score, while more appropriately distributed with the new UCB score.}
      \label{comparisonUCB}
\end{figure}

In summary, in the exploitation term of Eq. (\ref{eq6}), weighted average rather than pure average of the function values is calculated to earn an unbiased evaluation of the child subspace. And in the exploration term, the mean sampling density instead of the number of visits of subspaces is used to fully reflect the distribution of the historical sampling points in each child subspace. Moreover, by using a $log()$ function, the exploration term will be negative and the total $UCB_\rho$ will diminish when $\overline{\rho}_A < \overline{\rho}_B$, which means that the child subspace is explored even more thoroughly than the parent subspace. On the contrary, the exploration term will be positive when $\overline{\rho}_A > \overline{\rho}_B$ and be zero when $\overline{\rho}_A = \overline{\rho}_B$, which means that the child subspace is under-explored and equally explored compared to the parent subspace. By doing this, the algorithm is encouraged to sample more in the under-explored spaces to achieve a full coverage of the critical points in the whole search space. A comparison of the sampling frequency with the old and new UCB scores is shown in Fig. \ref{comparisonUCB}.

\subsection{Sampling with Advanced Local Samplers} \label{sample_with_others}
As mentioned before, for low-dimensional problems, random or Sobol algorithms with reject sampling are adequate to find potential critical points within a selected child search space. However, as the dimension of the search space increases, even a recursively partitioned subspace can be relatively large, in which random or Sobol sampling is no longer efficient. Therefore, advanced local samplers should be applied to guarantee the performance of the proposed algorithm in high-dimensional cases. 

In fact, with full coverage as the goal, LAMBDA acts as a guide to tell the local samplers which subspaces are valuable. Then the local samplers only need to focus on a certain subspace and find the optimal points, as they used to do in BBO problems. Note that the key of the adaptation is to make sure that the local samplers have a comprehensive understanding of the implicitly and arbitrarily bounded subspace. In this part, we illustrate how BO and TuRBO, two typical representatives of surrogate-based methods as advanced local samplers, are adapted with LAMBDA, especially how they uniformly generate candidates to be selected that fully cover the subspace.

\subsubsection{Sampling with BO}
\begin{figure}[t] 
      \centering
      \includegraphics[width=.95\linewidth]{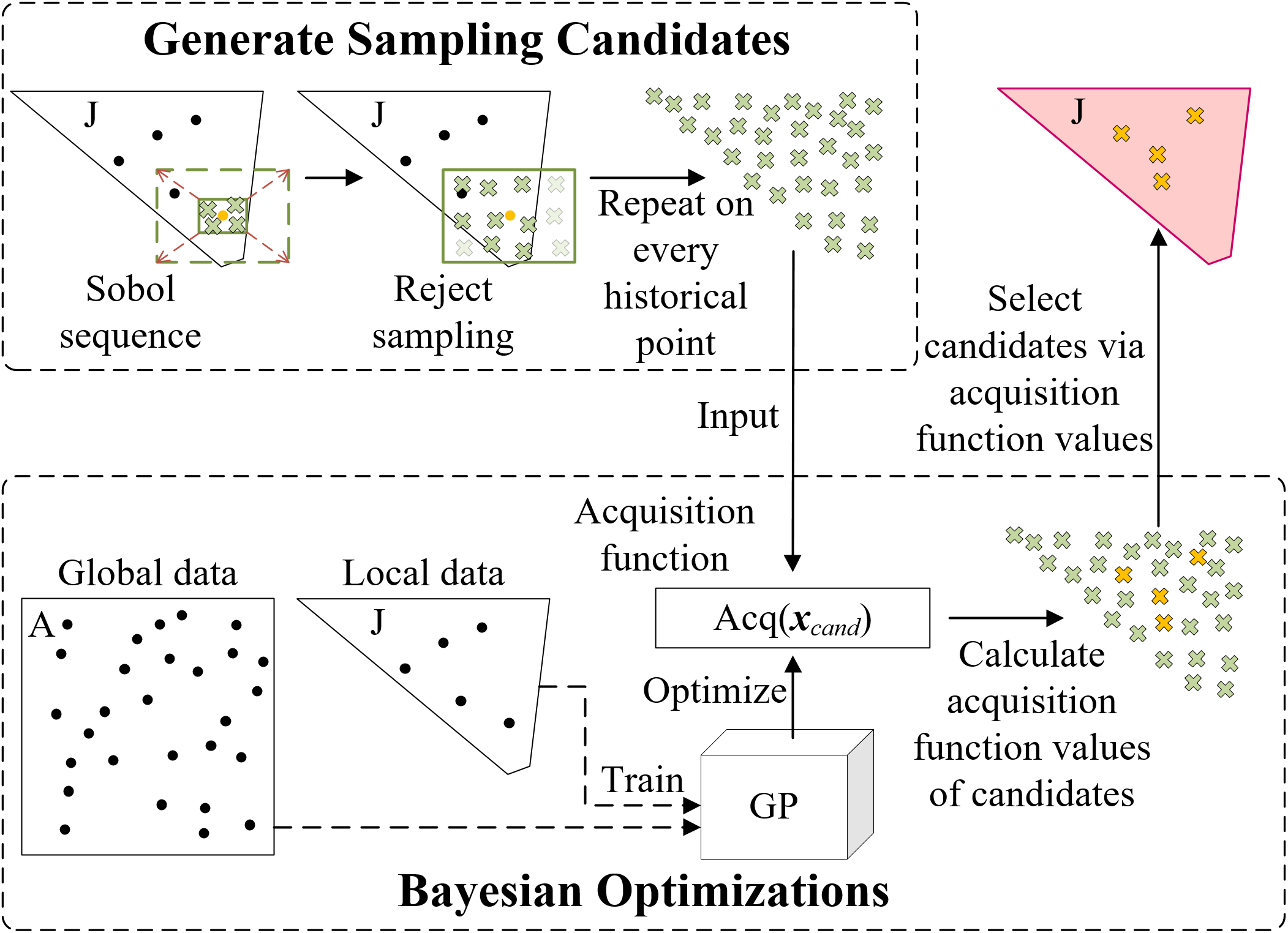}
      \caption{Illustration of sampling steps integrated with BO.}
      \label{sample_with_BO}
\end{figure}

BO utilizes Gaussian process (GP) models to select points with high acquisition function values from the candidate points for sampling. During this process, the quality of candidate points, namely coverage and uniformity, has a profound influence on BO. Generally, candidate points are randomly generated among the search space. However, since the boundaries of the partitioned subspace in LAMBDA are implicit and arbitrary, it is challenging to create points directly inside the subspace while guaranteeing quality. To solve this problem, we follow the idea proposed in \cite{wang2020learning}, detailed information can be found in Fig. \ref{sample_with_BO}.

As depicted in Fig. \ref{sample_with_BO}, we first generate the Sobol sequence around a certain historical point bounded by a (hyper)cube with an initial small length. Next, we expand the (hyper)cube recursively until there are Sobol points located outside the subspace. Reject sampling is then used to eliminate outliers. By repeating the above procedure on every historical point, a set of candidate points is obtained that are relatively uniformly distributed in the subspace, which will be the input of the acquisition function. In the second part of Fig. \ref{sample_with_BO}, GP models are trained by historical data either from the whole search space or from the selected subspace to optimize the acquisition function. After that, the function values of each candidate are calculated and the highest $n$ points are chosen to sample.

\subsubsection{Sampling with TuRBO}
\begin{figure}[t] 
      \centering
      \includegraphics[width=.95\linewidth]{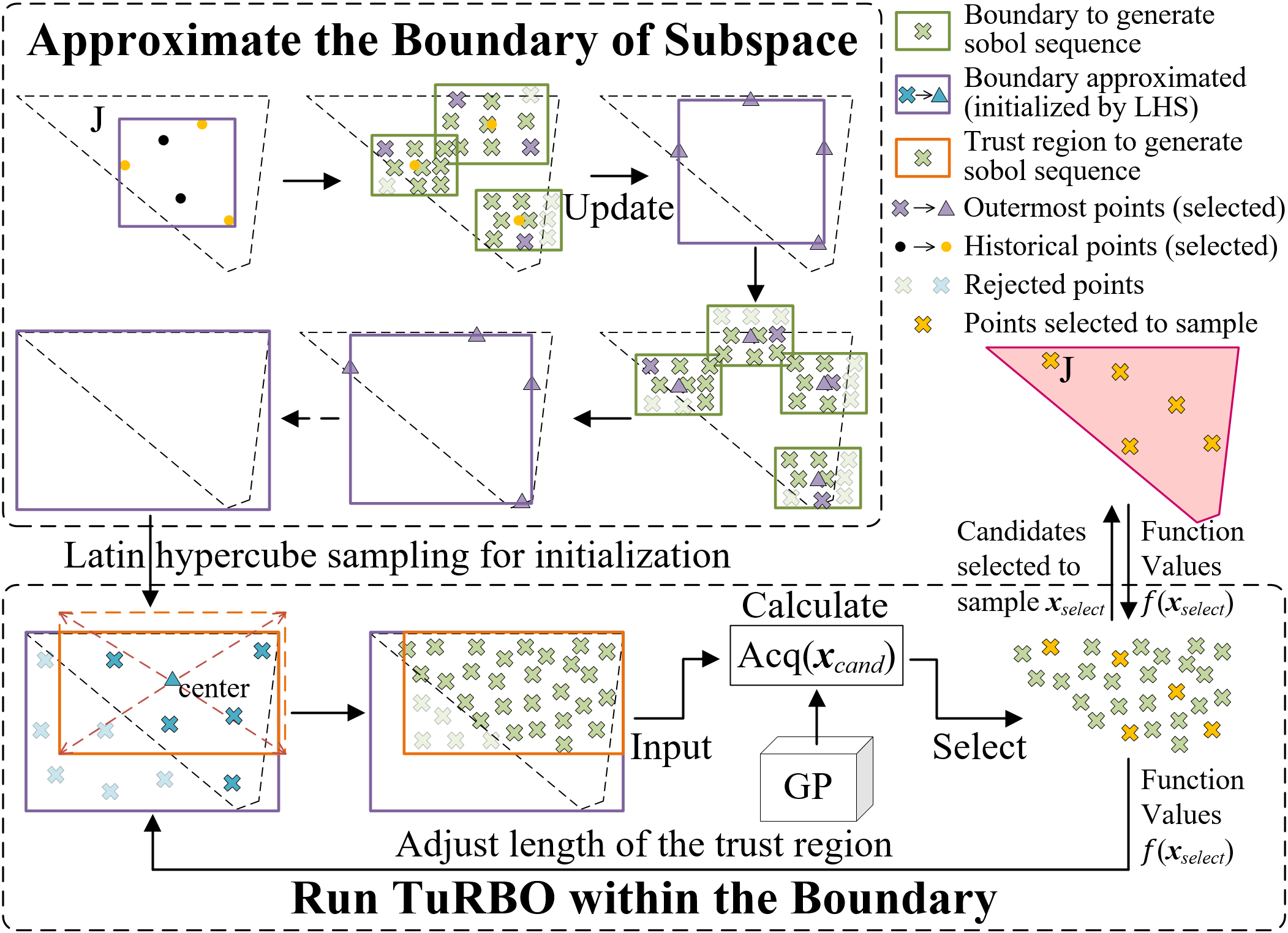}
      \caption{Illustration of sampling steps integrated with TuRBO.}
      \label{sample_with_TuRBO}
\end{figure}

Different from BO which samples the entire search space, TuRBO only samples a certain part of the search space called Trust Region (TR) and adjusts the length of TR dynamically according to the results of the last sample. Therefore, when applied with LAMBDA, TR should be established in the selected subspace. As the size of the subspace decreases with the search progress, it is difficult to predefine a fixed initial length for TR. To handle this issue, we first approximate an outer boundary for the subspace and then generate a TR relative to this boundary, e.g., the initial length of the TR is 0.8 times the length of the boundary on every dimension. Detailed information can be found in Fig. \ref{sample_with_TuRBO}.

As illustrated in Fig. \ref{sample_with_TuRBO}, in the upper part to approximate the boundary, we first obtain an initial boundary by finding the outermost points in the historical points. Then we generate a Sobol sequence as we have described before, but only around the outermost points. After reject sampling, an updated boundary with newly found outermost points is obtained. The above process will be repeated until no point in the subspace that can update the boundary is found. Finally, an outer boundary of the subspace is established, which will be used as the base for initializing TR. In the lower part of Fig. \ref{sample_with_TuRBO}, we initialize TuRBO-1 with a few samples by LHS which are generated within the approximated boundary and after reject sampling. Next, the point with the maximum or minimum value of the black-box function (depending on the purpose of the BBC problem) is chosen to be the center of TR. Then TR is initialized with this center and a length proportional to the length of the boundary. After that, we create the Sobol sequence as candidates $\boldsymbol{x}_{cand}$ within TR after reject sampling and calculate the acquisition function value of each candidate so as to select $\boldsymbol{x}_{select}$ to sample next. Finally, the black-box function values $f(\boldsymbol{x}_{select})$ are obtained and back-propagated to TuRBO so that it can adjust the length of TR adaptively and then decide whether to stop the search in current subspace by comparing the TR length with a predefined threshold.

\section{Experiments}
In this section, we first verify LAMBDA's performance and illustrate the advantages of LAMBDA over other baseline algorithms in terms of efficiency and coverage on synthetic functions. Baseline algorithms are chosen as mentioned in Section \uppercase\expandafter{\romannumeral2}, namely Random Search (RS) and Sobol , GA and DE, BO and TuRBO, and LaMCTS, which belong to random search algorithms, population-based algorithms, surrogate-based algorithms and MCTS-based algorithms, respectively. As for synthetic functions, we choose Holder-Table as the black-box function in low-dimensional BBC problems, while we can obtain a better visualization about the performance of each algorithm in low-dimensional cases. For high-dimensional BBC problems, we have designed a multimodal black-box function named Ripples (which we call a global optimum as a modality in this paper) so as to better demonstrate the superiority of our algorithm. Finally, a practical issue for the safety evaluation of certain ADSs based on SiL (Software-in-Loop) testing is demonstrated.
 
In addition, hyper-parameters of our algorithm and baselines are as demonstrated in Appendix \ref{Appendix_A}, noting that hyper-parameters of baselines are set according to the suggestion of their authors. If no suggestions are found, these hyper-parameters are set similarly to those of LAMBDA to ensure the fairness of comparisons. Furthermore, to avoid randomness, each experiment is repeated 10 times unless state otherwise, and each algorithm's mean performance is evaluated. Meanwhile, the number of points to evaluate the objective function is used to assess the efficiency of algorithms without bringing in confounders from platforms (e.g., OS) or hardware.

\subsection{Holder-Table Function Benchmarks}
\subsubsection{Test Function}
\begin{figure}[b] 
      \centering
      \includegraphics[width=.9\linewidth]{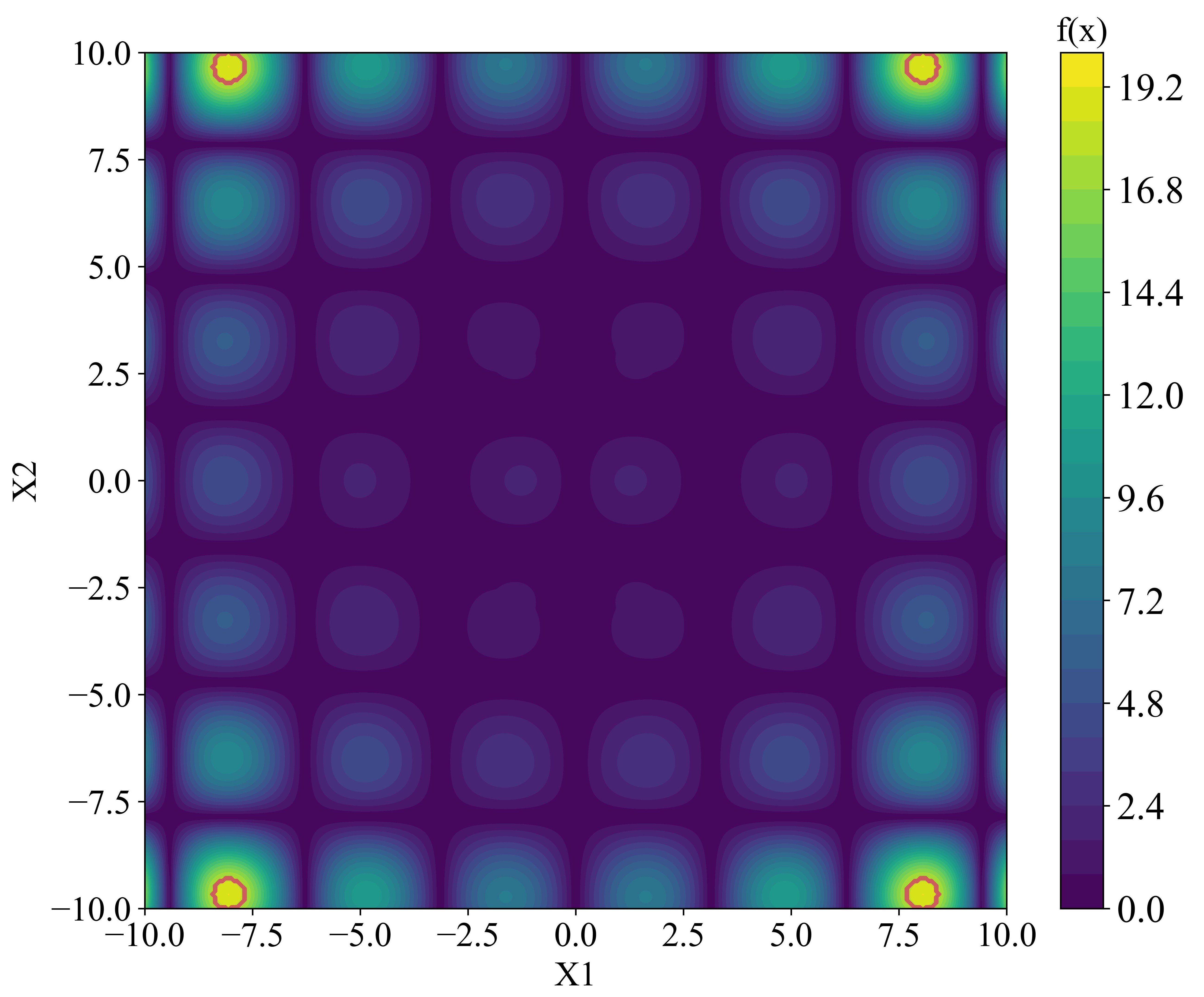}
      \caption{Holder-Table function.}
      \label{groundtruth_HT}
\end{figure}

Holder-Table is a classical 2-dimensional synthetic function to verify the performance of optimization algorithms. As shown in Eq. (\ref{eq10}) and Fig. \ref{groundtruth_HT}, it has 4 global optima distributed at the corners of the search space, namely $f(\boldsymbol{x}^*) = 19.2085$, at $\boldsymbol{x}^* = (8.05502,9.66459)$, $(8.05502,-9.66459)$, $(-8.05502,9.66459$), and $(-8.05502,-9.66459)$, while many local maxima to trap the optimizer. It should be noted that in this paper we invert the function by multiplying $-1$ by the original function, since LAMBDA is a maximization algorithm. The black-box inequality is defined as $f(\boldsymbol{x}) > 18$, whose solution set is lined with red. Moreover, as described above, LAMBDA uses Sobol as the local sampler for this low-dimensional case, and so does LaMCTS for fair comparison.

\begin{equation}\label{eq10}
\begin{aligned}
f(\boldsymbol{x})=\left|sin(x_1)cos(x_2)\text{exp}\left(\left|1-\frac{\sqrt{x_1^2+x_2^2}}{\pi}\right|\right)\right| \\ \ x_i \in [-10,10] \ \text{for all}\  i = 1, 2
\end{aligned}
\end{equation}

\subsubsection{Experimental Process and Results}
\begin{figure}[t] 
      \centering
      \includegraphics[width=\linewidth]{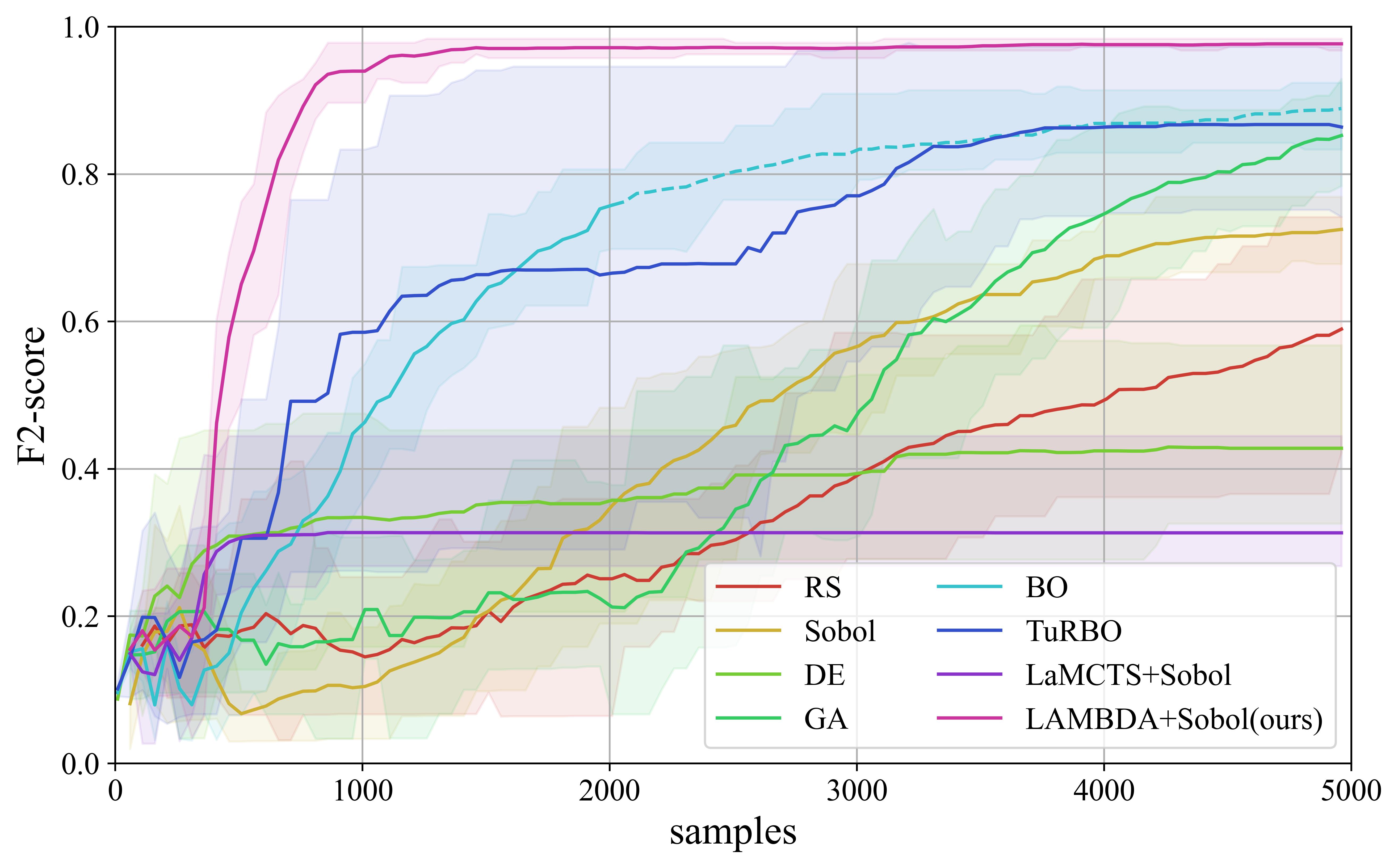}
      \caption{Benchmarks on Holder-Table function.}
      \label{benchmark_HT}
\end{figure}

After every 10 samples, the F2 score of each algorithm is calculated under current historical sampling records as described in Section \ref{metric}. With the search progressing, the F2 score will gradually increase and eventually reach 1 if all four modalities are covered. The benchmark results are shown in Fig. \ref{benchmark_HT}. The solid lines represent the average F2 scores of algorithms while the light-colored areas lie between test cases with the algorithms' best and worst performances, indicating the stability of the algorithms. What is more, it is worth noting that BO (with GP) suffers from the complexity of $O(n^3)$, therefore, it is hard to accept the complexity of BO when the number of sample records increases. Specifically, in this experiment, BO needs more than 60s to suggest the next batch of sampling points when the sample records exceed 2000. As a result, although we still run and evaluate BO after 2000 samples (dashed line in Fig. \ref{benchmark_HT}), it is no longer meaningful in terms of time cost.

As can be seen from Fig. \ref{benchmark_HT}, LAMBDA achieves the best performance among all algorithms in terms of efficiency (rate of increase of the F2 score), coverage (final F2 score), and stability (range of the light-colored area). The F2 score of LAMBDA has stabilized above 0.95 in only 1500 iterations of sampling, achieving a speedup of 33x when compared with RS, which needs to run around 50000 samples to reach the same level of F2 score. Among all baselines, TuRBO performs the best. In the best of the ten trails, TuRBO obtains a similar F2 score and convergence rate with the worst performance of LAMBDA. But the large variance has dragged down TuRBO’s mean performance. As the predecessor of LAMBDA, LaMCTS does not work well on this problem since it has been trapped in one of the four modalities due to the sampling bias. Thus, the F2 score always converges to about 0.25. Besides, Sobol performs better than RS, indicating that sampling uniformity plays an important role in getting more coverage and further proving the advantage of using the Sobol sequence for initialization in our algorithm.

\begin{figure*}[t] 
      \centering
      \includegraphics[width=.95\linewidth]{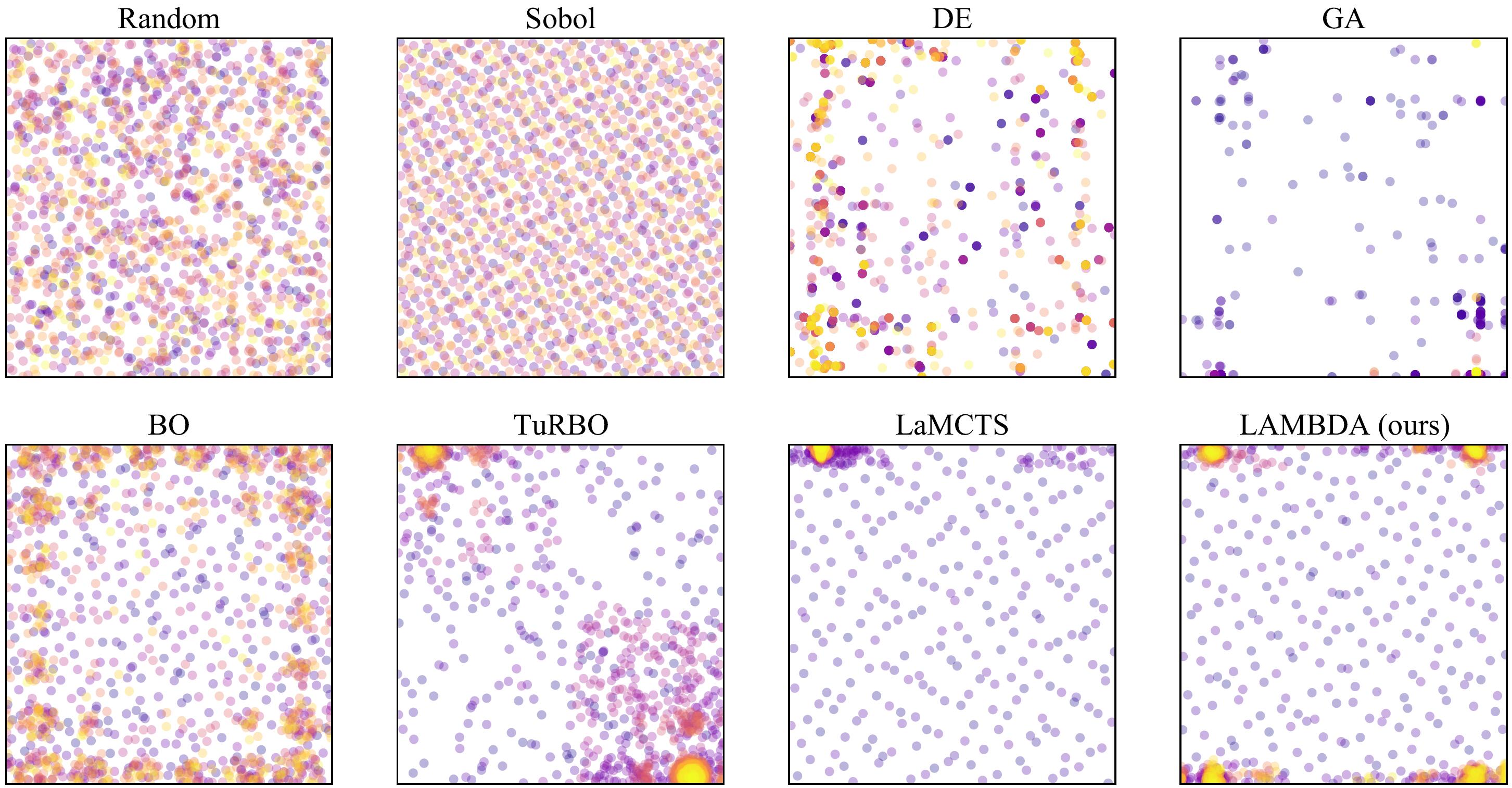}
      \caption{Sampling dynamics of each algorithm with 1500 samples. Color from purple to golden represents the record being sampled at the beginning to the end of the optimization.}
      \label{dynamic_HT}
\end{figure*}

To better visualize the performance of each algorithm, the results of sampling dynamics are shown in Fig. \ref{dynamic_HT}. The sampling of LAMBDA, BO and DE is relatively global, covering the directions of 4 key modalities in the search space. However, BO and DE are distracted into the local maxima during sampling, resulting in a waste of sampling budgets. As for other baselines, both TuRBO and GA only find 2 of the 4 modalities, while LaMCTS only finds 1 modality in the upper left corner. When it comes to random search algorithms, compared with RS, the sampling distribution of Sobol is more uniform, and the sampling density of Sobol is almost the same in search space. To sum up, the sample points of the baseline algorithms with lower F2 scores tend to gather in certain small areas. They only try some global exploration of the search space in the early stage of optimization, and then no longer jump out of the currently focused modality for sampling.

\subsection{Ripples Function Benchmarks}
\subsubsection{Test Function}
To further validate the performance of LAMBDA, high-dimensional experiments are required. Recall that the purpose of our experiments is to test the algorithm's coverage of critical subspaces, high-dimensional synthetic functions with multiple global optimums are preferred. However, to the best of the authors' knowledge, there exist few functions satisfying the above requirement. Therefore, in this paper, we have designed a high-dimensional multimodal synthetic function named Ripples that specifically serves as a black-box objective function for the BBC problem. The formula of Ripples is shown in Eq. (\ref{eq11}).

\begin{equation}\label{eq11}
\begin{aligned}
f(\boldsymbol{x})= \sum _{i = 1} ^ {dim} \left(e^{-\frac{x_i^2}{2\sigma^2}} + k cos(\omega x_i) - k \right), \ \text{where} \ x_i = \\ \Vert \boldsymbol{x} + M_{i,:} \Vert_2, \ \text{for all} \  i = 1, 2, ..., dim
\end{aligned}
\end{equation}
In which, 

\begin{equation}\label{eq12}
M  = I_{dim} \times bias
\end{equation}
where $dim$ is the dimension of the function. $I_{dim}$ is an identity matrix with order $dim$. $M_{i,:}$ represents the $i$-th row of the matrix $M$. $\sigma$, $k$, $\omega$ and $bias$ are parameters that determine the shape of the function and the locations of the optimal points.

As depicted in Eq. (\ref{eq11}), there are two parts inside the $\Sigma()$ function. The former part is actually a Gaussian function with a default $\mu = 0$. The latter part is a $cos()$ function to generate multiple local maxima, which shares the same idea with the classical Rastrigin function. It should be noted that $\omega$ in the $cos()$ function must be $\sqrt{2}$ or a multiple of $\sqrt{2}$ to ensure that the global optimums appear at the expected positions. With the above settings, there are a total of $dim$ modalities in the $dim$-dimensional Ripples, namely $f(\boldsymbol{x}^*) = 1$, at $\boldsymbol{x}_1 = (-bias, 0, 0, ..., 0)$, $\boldsymbol{x}_2 = (0, -bias, 0, ..., 0)$, $\boldsymbol{x}_3 = (0,0,-bias, ..., 0)$, $...$, $\boldsymbol{x}_{dim} = (0, 0, 0, ..., -bias)$, respectively.

\begin{figure*}[t]%
    \centering
    \subfloat[One-dimensional Ripples]{
        \label{1d_Ripples}
        \includegraphics[width=0.285\linewidth]{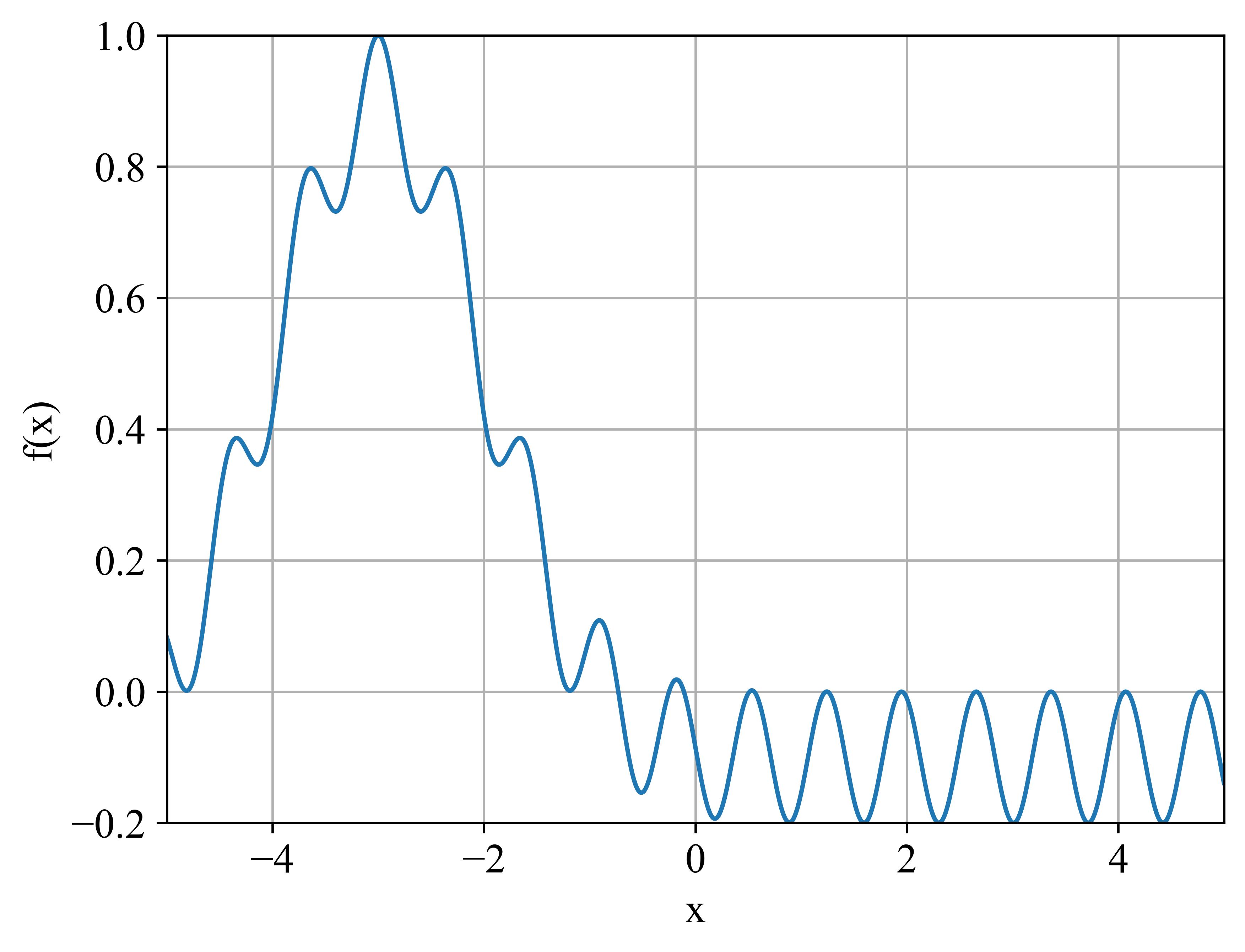}
        }\hfill
    \subfloat[Two-dimensional Ripples]{
        \label{2d_Ripples}
        \includegraphics[width=0.275\linewidth]{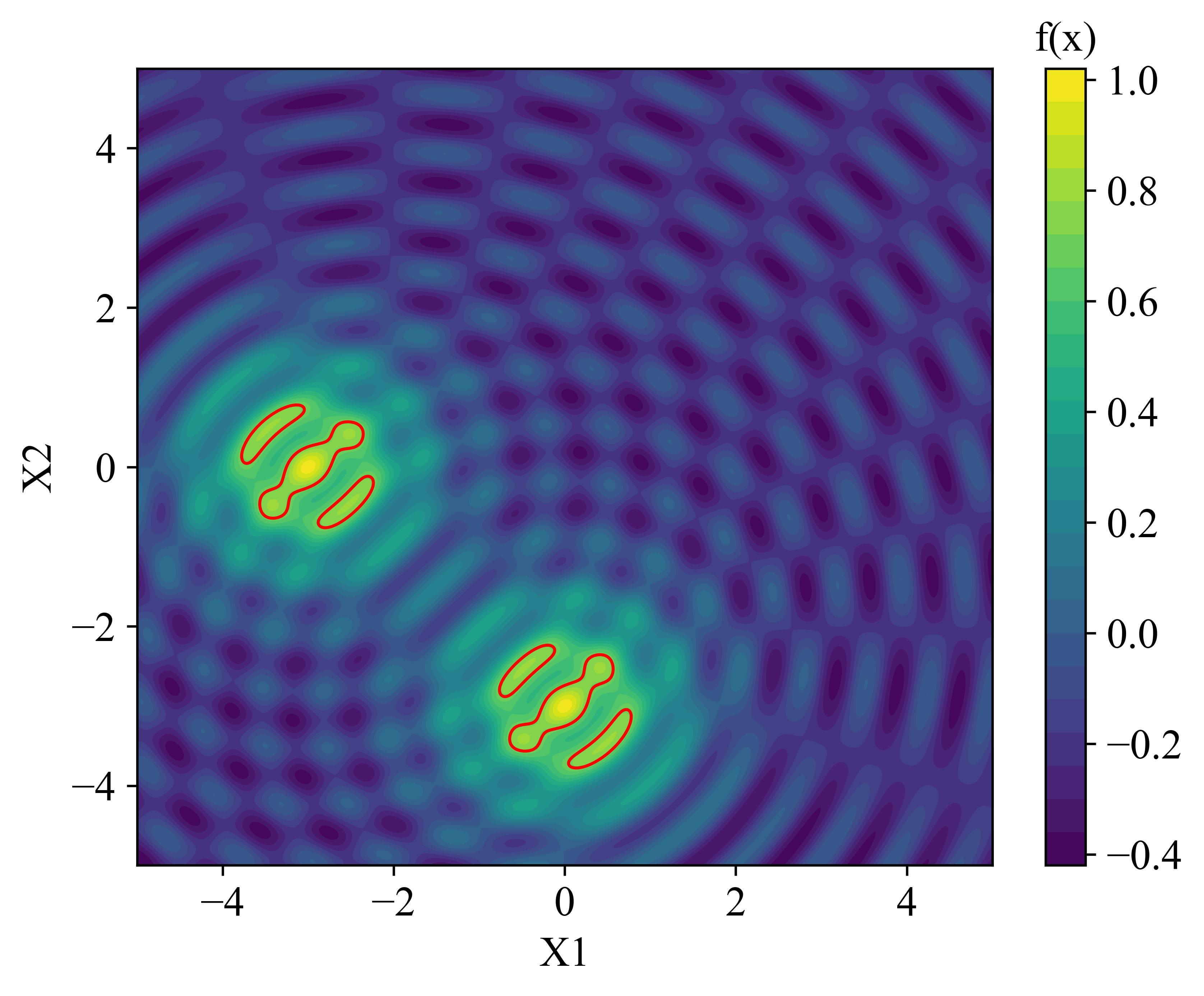}
        }\hfill
    \subfloat[Three-dimensional Ripples]{
        \label{3d_Ripples}
        \includegraphics[width=0.35\linewidth]{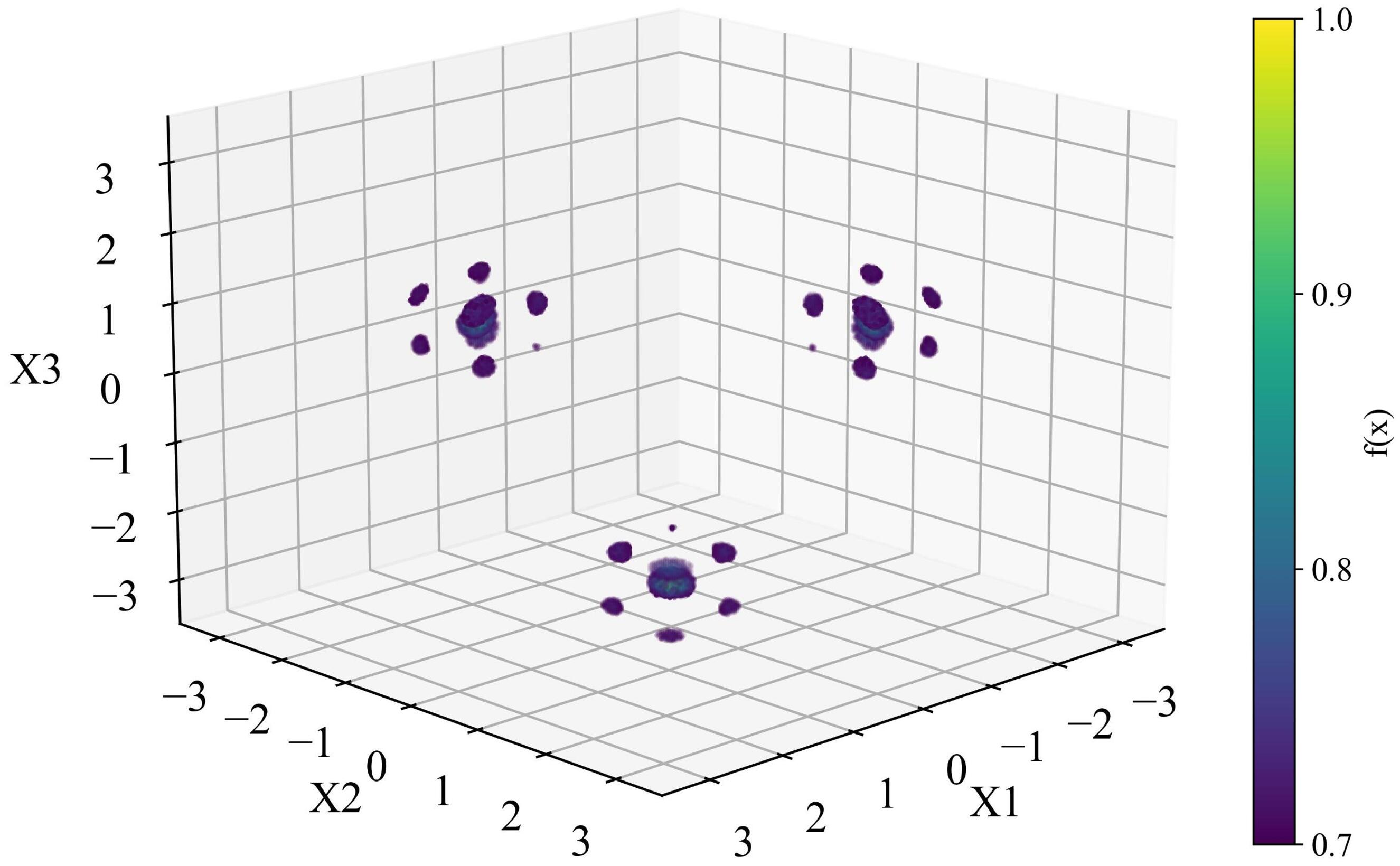}
        }
    \caption{Illustration of Ripples function.}
    \label{Ripples}
\end{figure*}

Specifically in this paper, we set $bias = 3$, $\sigma = 1$, $\omega = 2\sqrt{2}$ and $k = 0.1$. In addition, the input domain is chosen to be $[-5,5]$ in every dimension. Fig. \ref{Ripples} demonstrates the profiles of Ripples under the dimension of 1 to 3, respectively. As described above, in Fig. \ref{1d_Ripples}, there is only one modality located at $x = -3$ in 1-d Ripples. While in 2-d Ripples, two modals are obtained with locations at $(-3, 0)$ and $(0, -3)$. Moreover, the solution sets that satisfy $f(\boldsymbol{x}) > 0.7$ are lined by red lines in Fig. \ref{2d_Ripples}. As for 3d-Ripples in Fig. \ref{3d_Ripples}, we only illustrate the points whose function values are higher than 0.7 to obtain a clearer visualization result. And the three modalities right appear at the expected locations, namely $(-3,0,0)$, $(0,-3,0)$ and $(0,0,-3)$. From the above observations, we can conclude that Ripples perfectly satisfies our requirements with multiple global optimums independently distributed at known positions, as well as a low percentage of critical areas and a large number of local maxima that are challenging enough for optimization algorithms.

\subsubsection{Experimental Process and Results}
\begin{figure}[h]%
    \centering
    \subfloat[Benchmarks on optimization algorithms]{
        \label{benchmark_5dRip}
        \includegraphics[width=\linewidth]{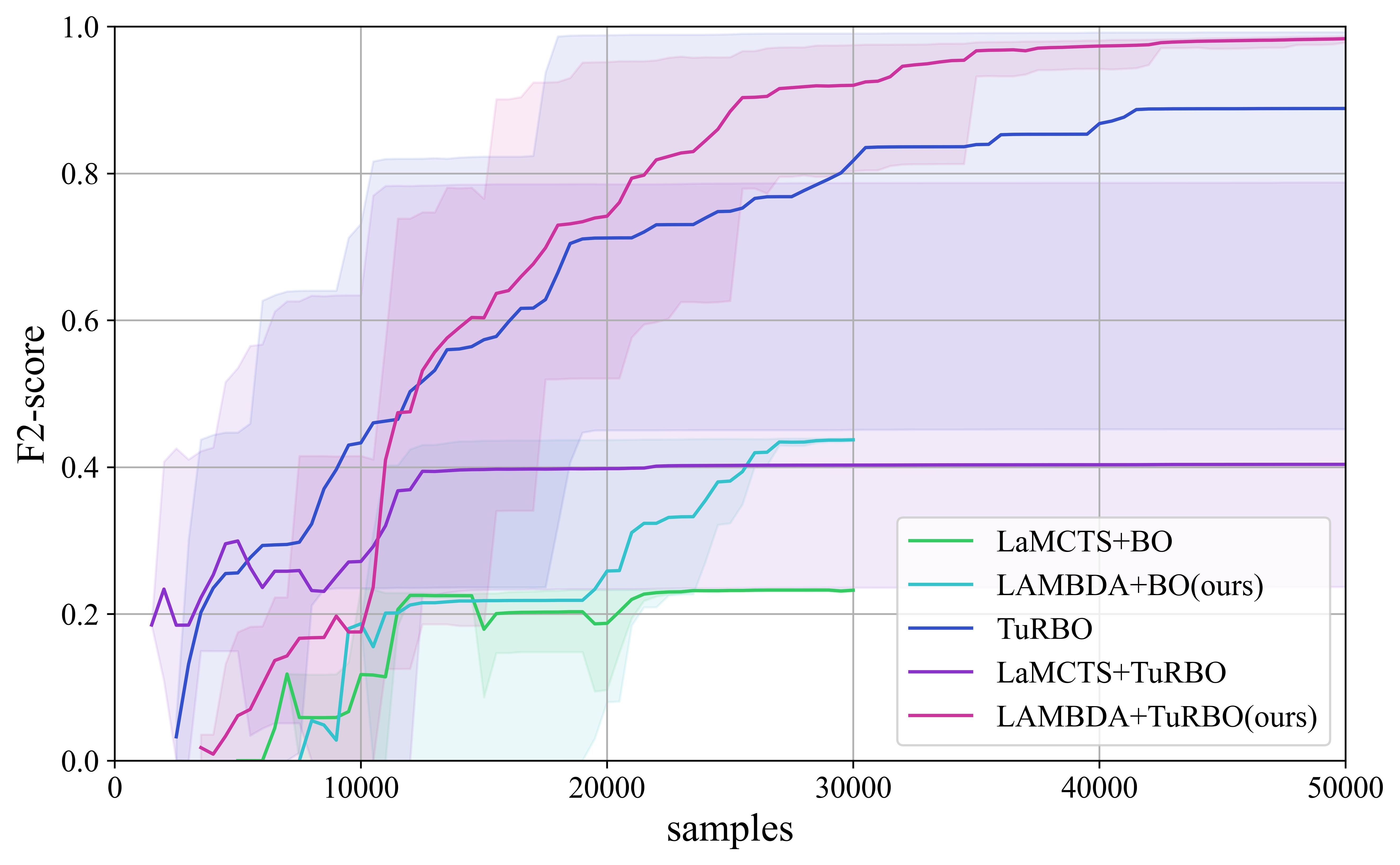}
        }\hfill
    \subfloat[Comparative experiments using RS and Sobol]{
        \label{benchmark_MC_Sobol}
        \includegraphics[width=.98\linewidth]{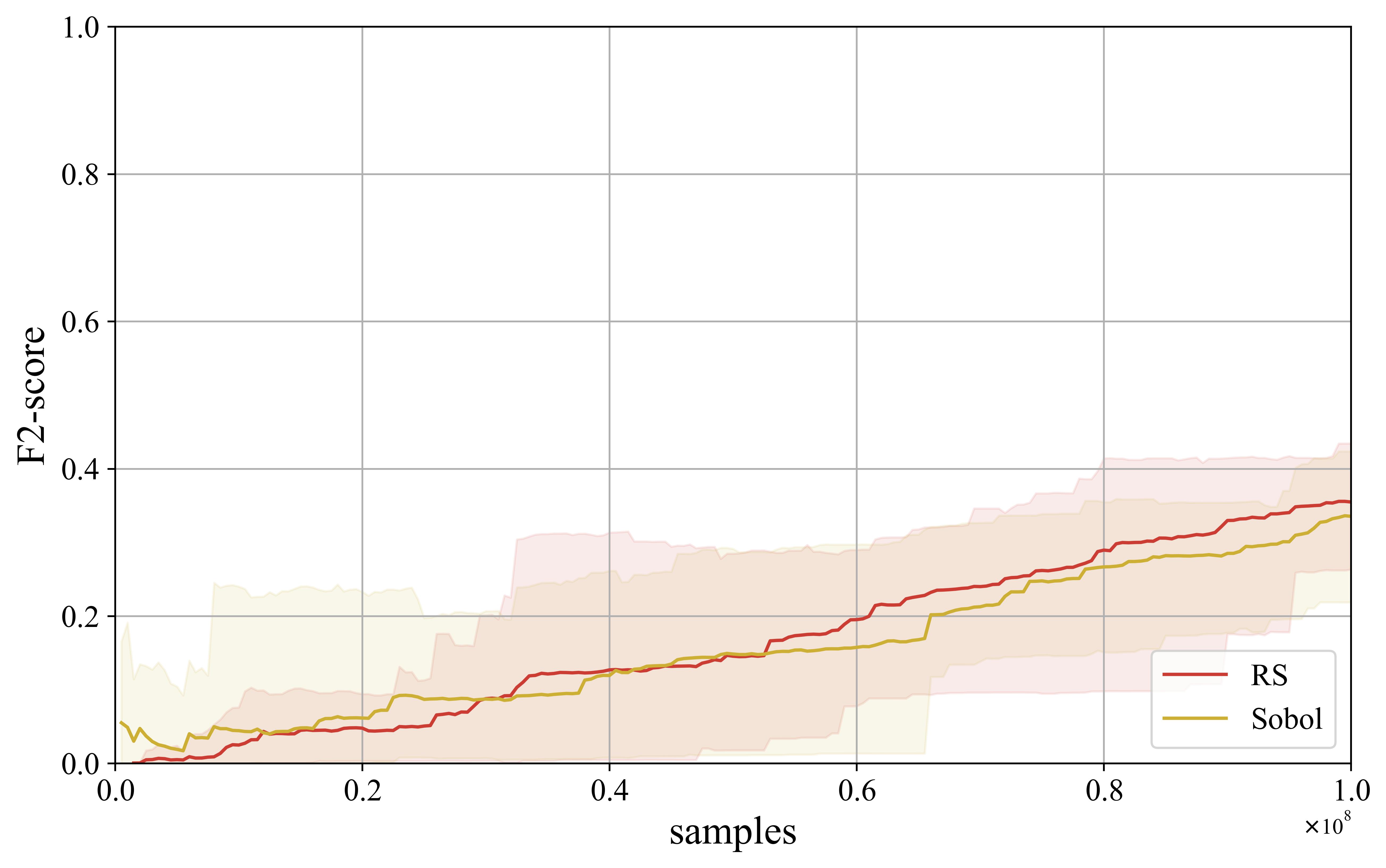}
        }
    \caption{Benchmarks on 5-dimensional Ripples function.}
    \label{5d-Ripples}
\end{figure}

In the experiment, we benchmark the algorithms with a 5-dimensional Ripples using the parameters mentioned above. The black-box inequality is defined as $f(\boldsymbol{x}) > 0.7$. 
Considering the high-dimensional feature, we use BO and TuRBO as local samplers for LaMCTS and LAMBDA, respectively. The experimental results are shown in Fig. \ref{benchmark_5dRip}. Unfortunately, neither random search algorithms (RS and Sobol) nor population-based algorithms (DE and GA) can find even a single modality within 50,000 sampling points. And BO (with GP) still suffers from the complexity of $O(n^3)$ to the extent that it cannot complete the sampling. Moreover, with BO as the local sampler, both LaMCTS and LAMBDA are time-costing, resulting in only 30,000 samples and 3 repetitions of the experiment. When it comes to the other three algorithms in Fig. \ref{benchmark_5dRip}, LAMBDA with TuRBO performs best, which ends up with an average F2 score of 0.98, meaning that in each of the ten trails, LAMBDA can always cover the five modalities. TuRBO rises the fastest within the first 10,000 samples, but is then overtaken by LAMBDA, with an average F2 score of 0.89 in the end. In particular, it can be found that the best case of TuRBO achieves the full coverage of all modalities even faster than LAMBDA, but the worst case only obtains an F2 score of 0.45. This feature will be discussed later. Similarly to that in the low-dimensional case, the F2 score of LaMCTS never exceeds 0.8 due to sampling bias and remains at an average of 0.4.
To further quantify the superiority of our algorithms, comparative experiments using RS and Sobol with a sampling budget of 100 million points are conducted. As the results shown in Fig. \ref{benchmark_MC_Sobol}, it can be estimated that our algorithm achieves a speedup of 6000x when compared with RS and Sobol, which need to run around $3\times10^{8}$ samples to reach the same level of F2 score. 

\begin{figure*}[t]%
    \centering
    \subfloat[The best case of TuRBO]{
        \label{turbo_best}
        \includegraphics[width=0.3\linewidth]{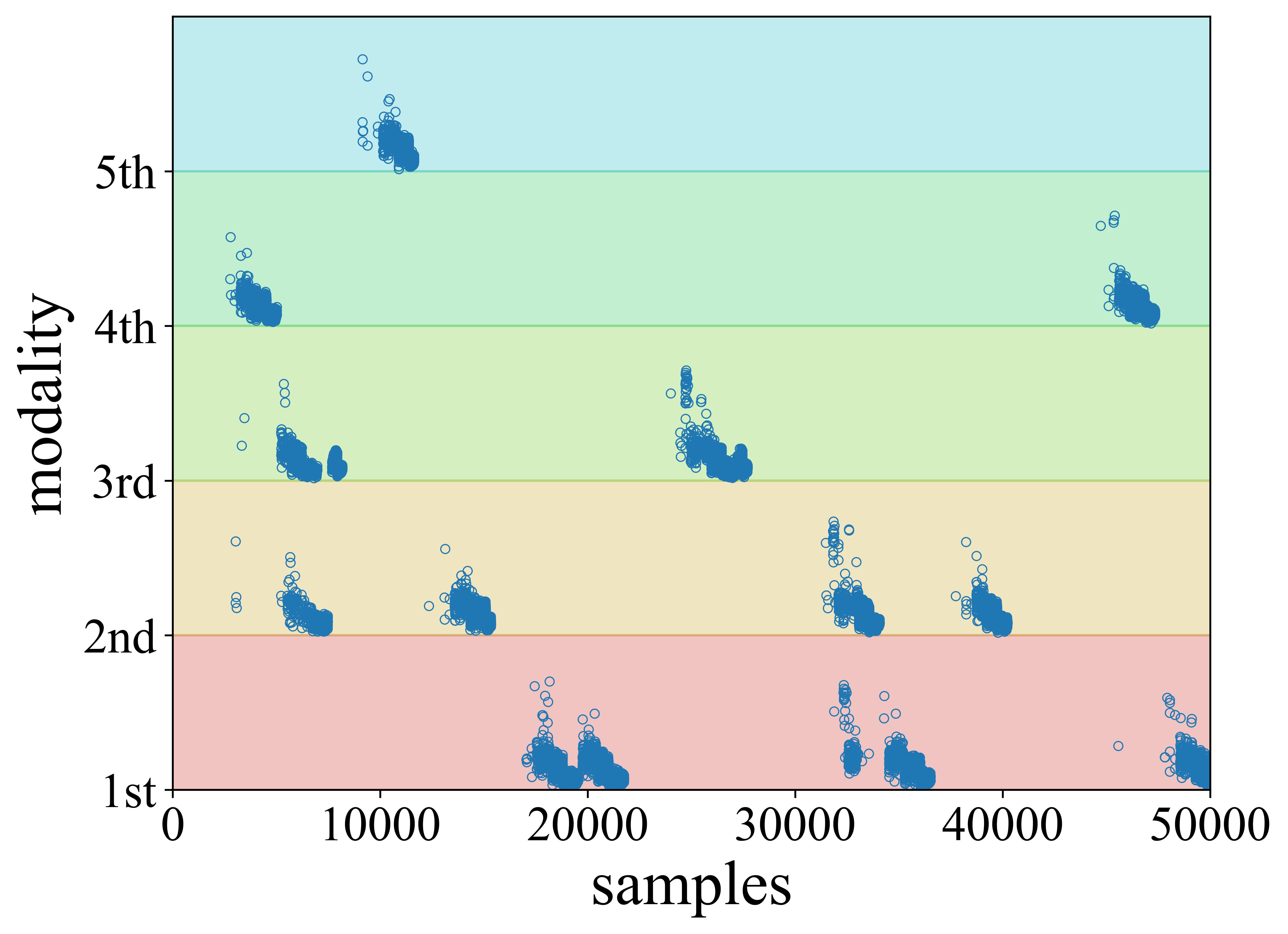}
        }\hfill
    \subfloat[The best case of LaMCTS (with TuRBO)]{
        \label{lamcts_best}
        \includegraphics[width=0.3\linewidth]{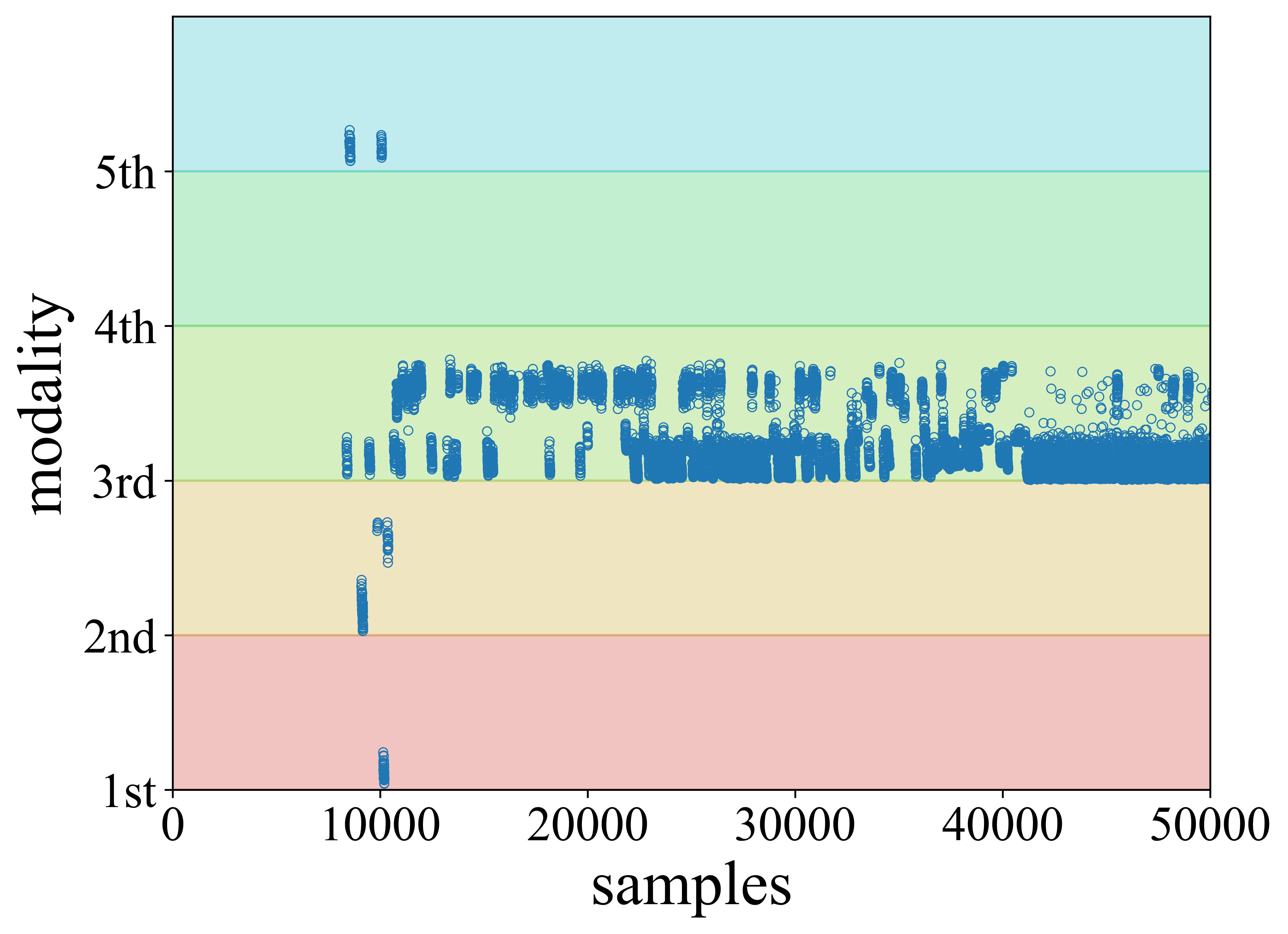}
        }\hfill
    \subfloat[The best case of LAMBDA (with TuRBO)]{
        \label{lambda_best}
        \includegraphics[width=0.3\linewidth]{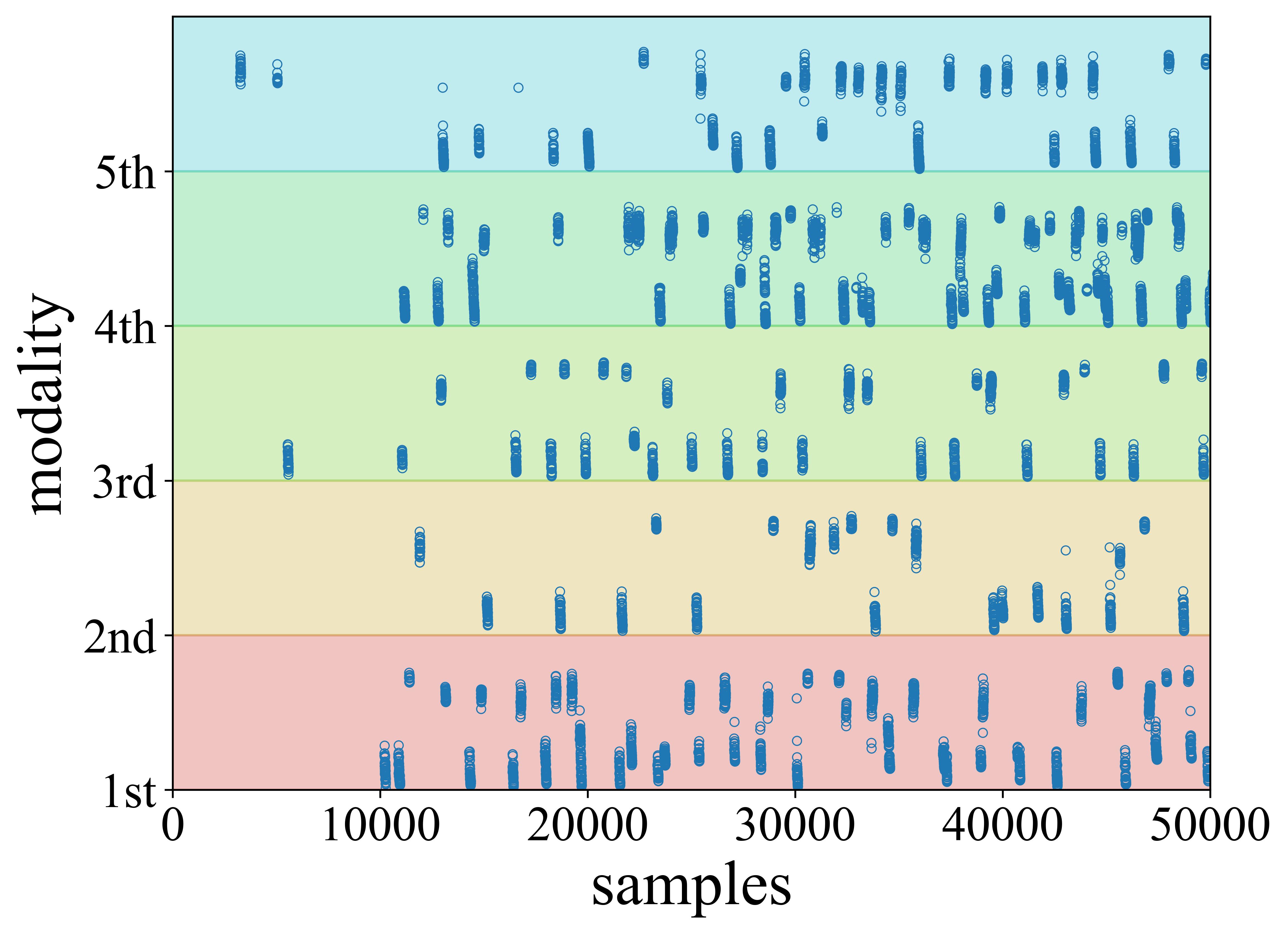}
        }
    \newline
    \subfloat[The worst case of TuRBO]{
        \label{turbo_worst}
        \includegraphics[width=0.3\linewidth]{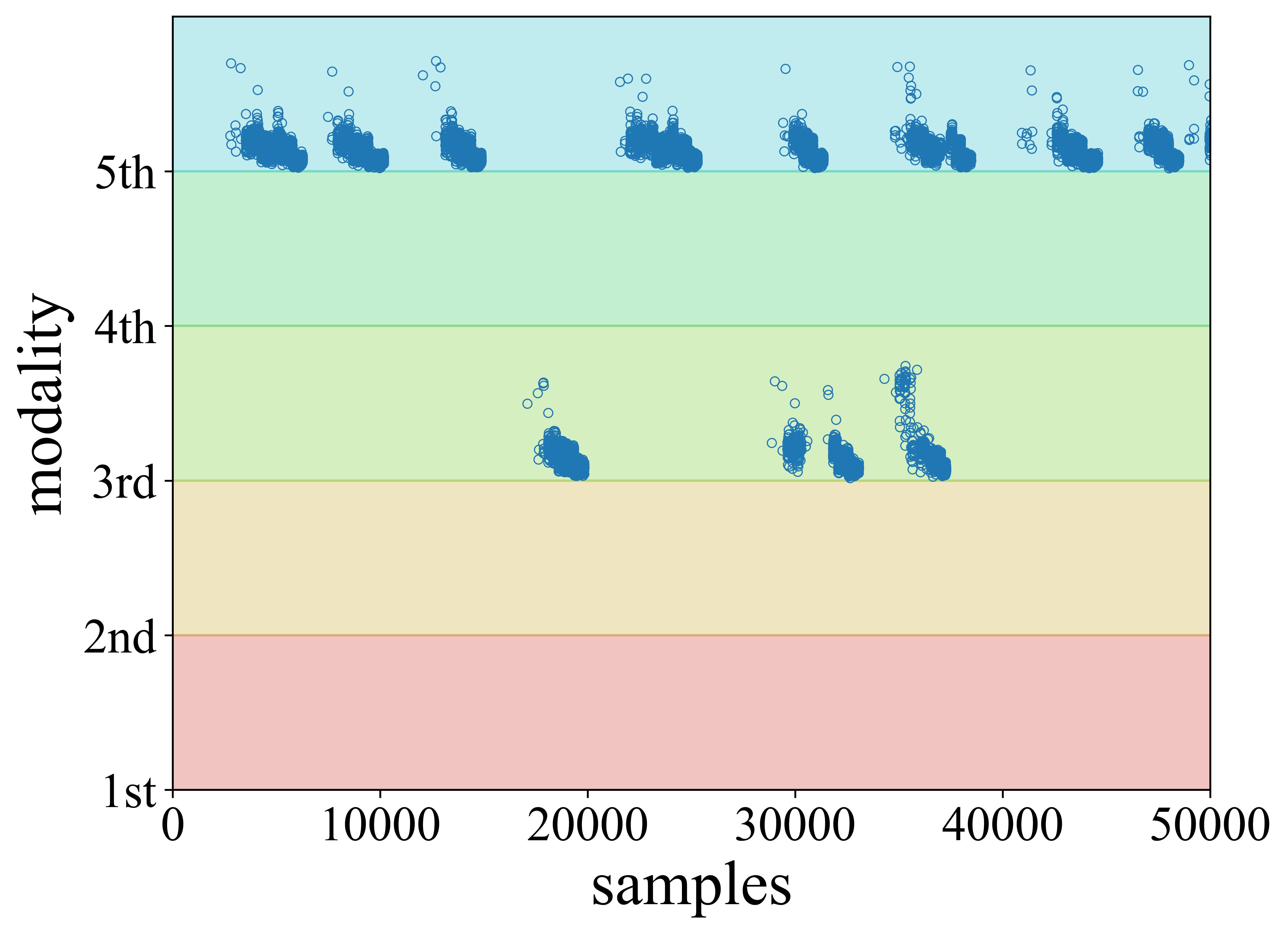}
        }\hfill
    \subfloat[The worst case of LaMCTS (with TuRBO)]{
        \label{lamcts_worst}
        \includegraphics[width=0.3\linewidth]{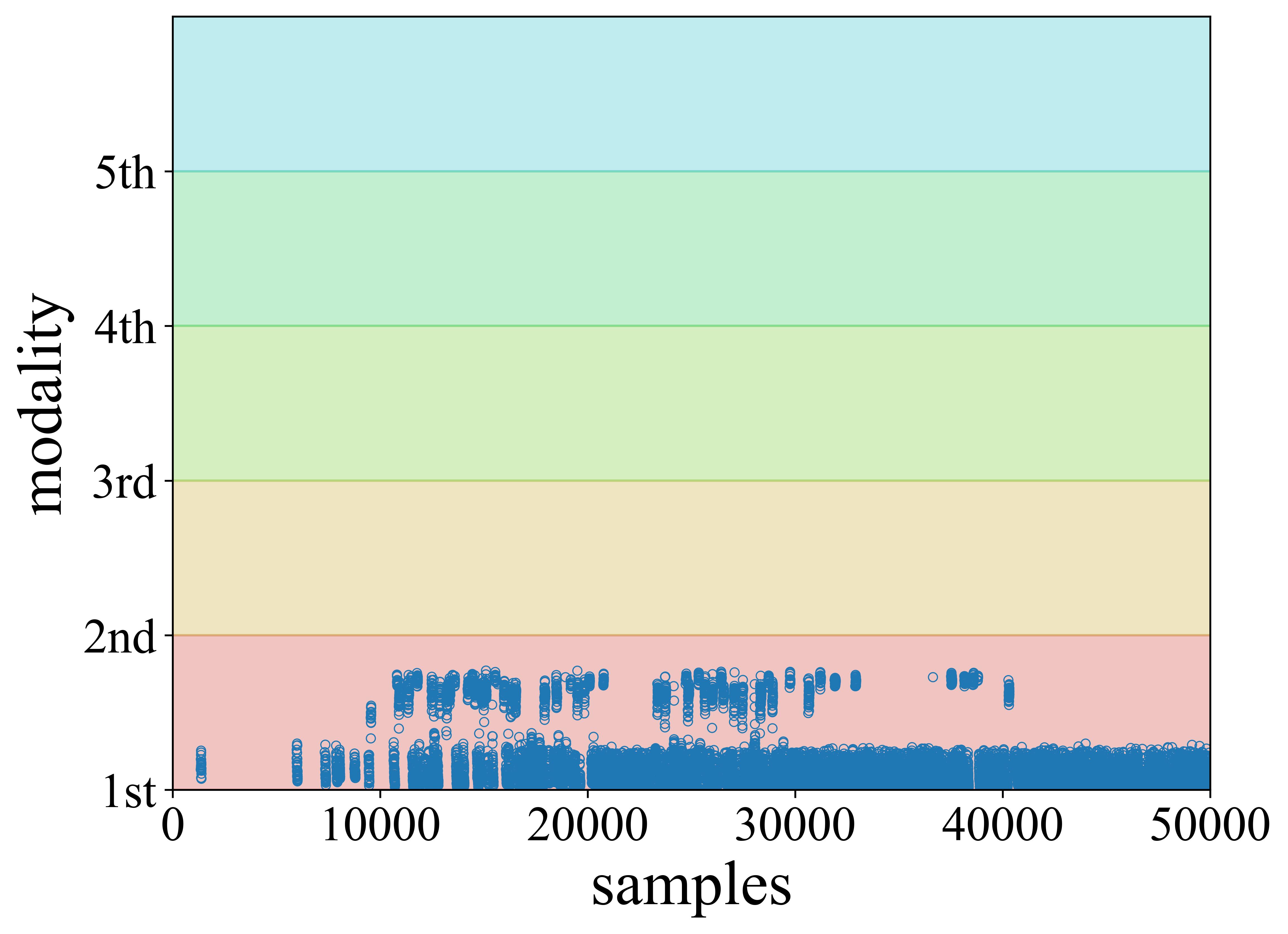}
        }\hfill
    \subfloat[The worst case of LAMBDA (with TuRBO)]{
        \label{lambda_worst}
        \includegraphics[width=0.3\linewidth]{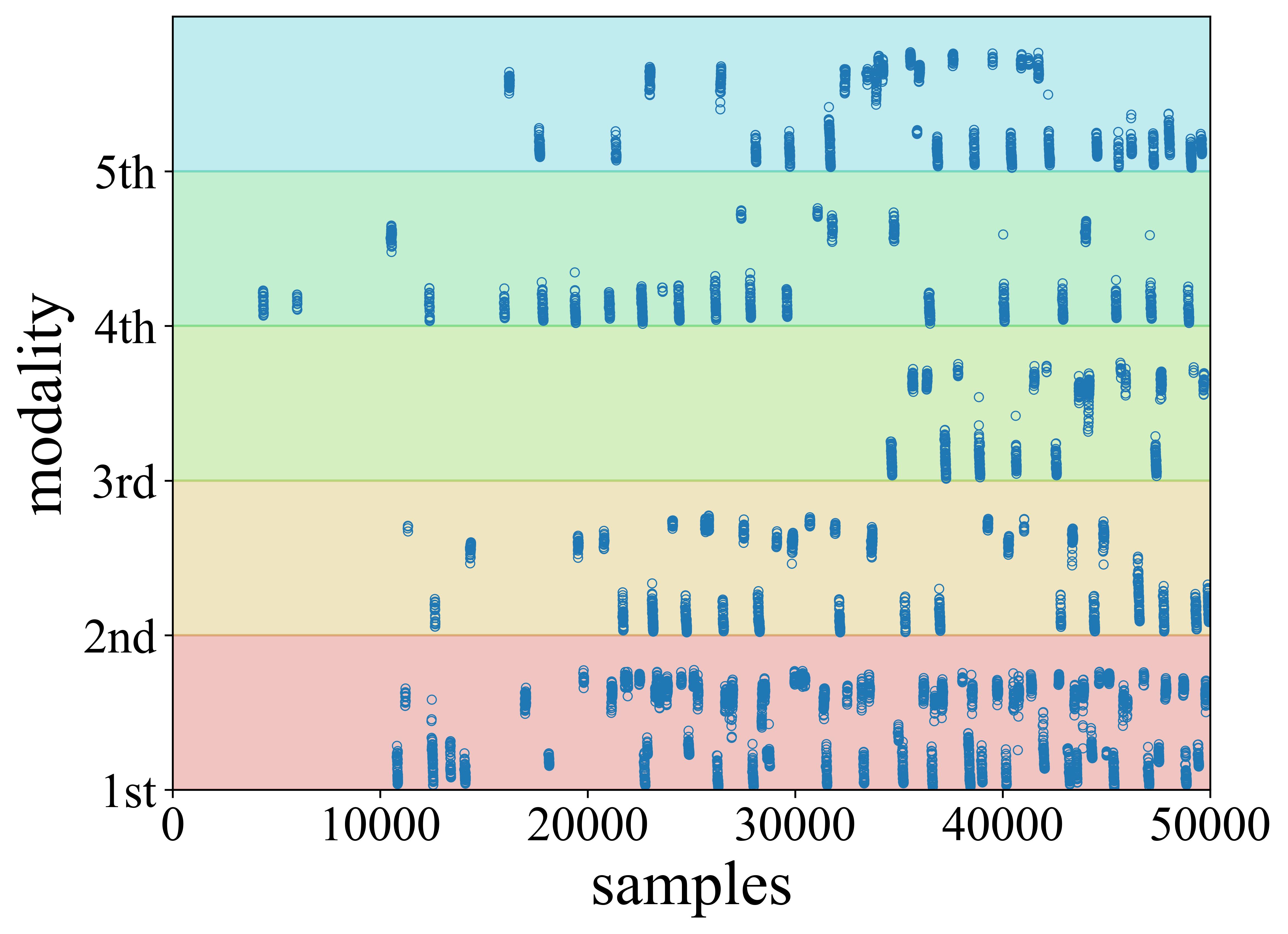}
        }
    \caption{Sampling dynamics of TuRBO, LaMCTS and LAMBDA within 50,000 samples, with the cases of their best and worst performance.}
    \label{ModalDistribution}
\end{figure*}

For this high-dimensional case, apparently it is hard to directly visualize the distribution of the sampling points. As a remedy, we first select the sampling points that satisfy the black-box inequality, and then calculate the distance between the selected sampling point and its modality center. Since the function values of the selected sampling points are all greater than 0.7, it is easy to determine which modality the sampling point belongs to (the distance between the sampling point and its own modality is significantly smaller than the distance to the other modalities). The results are illustrated in Fig. \ref{ModalDistribution}. Here, we give the results of the best and worst performance of the algorithms, respectively. Taking Fig. \ref{turbo_best} as an example, each bar with a specific color represents one modality, and the distance between the blue point and the bottom line of the bar represents the distance between the sampling point and the modality center. In this way, the sampling dynamics is well demonstrated and the sampling characteristics of the different algorithms are clearly distinguished.

From Fig. \ref{ModalDistribution}, we can find that the solution sets of the black-box inequality (namely, the critical points) found by TuRBO are distributed in clusters, whereas the critical points found by the MCTS-based methods are distributed in bars. The reason for this discrepancy is that TuRBO is based on multiple simultaneously initialized TRs, each of which searches globally without interfering with each other and ends up sampling only one modality. In contrast, the MCTS-based methods are based on the simultaneous search of multiple partitioned subspaces, each of which is allocated only a limited sampling budget. Fig. \ref{lamcts_best} and Fig. \ref{lamcts_worst} give us intuitive examples of how sampling bias affects LaMCTS. Even in the best case, LaMCTS still focuses on only one modality in the end, despite the fact that critical points in other modalities have been found in the beginning. 

When comparing the best cases of TuRBO and LAMBDA, TuRBO outperforms LAMBDA in terms of the speed of finding all modalities and the number of critical points found, leading to a faster rise in F2 scores in Fig. \ref{benchmark_5dRip}. However, as mentioned above, each TR initialized by TuRBO searches independently and finds a modality randomly without interfering with each other. The drawback of this mechanism is that it cannot guarantee that the modalities randomly found by different TRs will cover all modalities exactly. This limitation can be proved in Fig. \ref{turbo_worst}, where the number of critical point clusters is almost the same as it is in Fig. \ref{turbo_best}, implying that the search efficiency is the same, but the searched points are concentrated in the third and fifth modality, resulting in an F2 score of only 0.45. As a comparison, LAMBDA ensures full coverage of all modalities by selecting multiple subspaces to search at once using Beam Search and encouraging the search for unexplored regions through UCB scores. As can be seen in Fig. \ref{lambda_worst}, although LAMBDA loses the third modality in the first 35,000 samples, the relatively high UCB scores of the subspace around the third modality eventually motivate LAMBDA to search towards this region. To sum up, besides the average efficiency and coverage, the stability (or call it confidence level) of the optimization algorithms is also very important, especially in the safety evaluation of autonomous vehicles, where we cannot know the total critical modality number in advance.

\subsection{Practical Issue for Safety Evaluation of ADSs}
The above two experiments have well verified the superiority of LAMBDA over other baseline algorithms on low- and high-dimensional synthetic functions, respectively. For practical applications, safety evaluation of an ADS in a particular logical scenario \cite{menzel2018scenarios} is a typical BBC problem. Here, we give a practical example for safety evaluation of a certain ADS through SiL testing based on Virtual Test Drive (VTD) \cite{hexagon2022}. 

\subsubsection{Test Scenario}

\begin{figure}[b] 
      \centering
      \includegraphics[width=.9\linewidth]{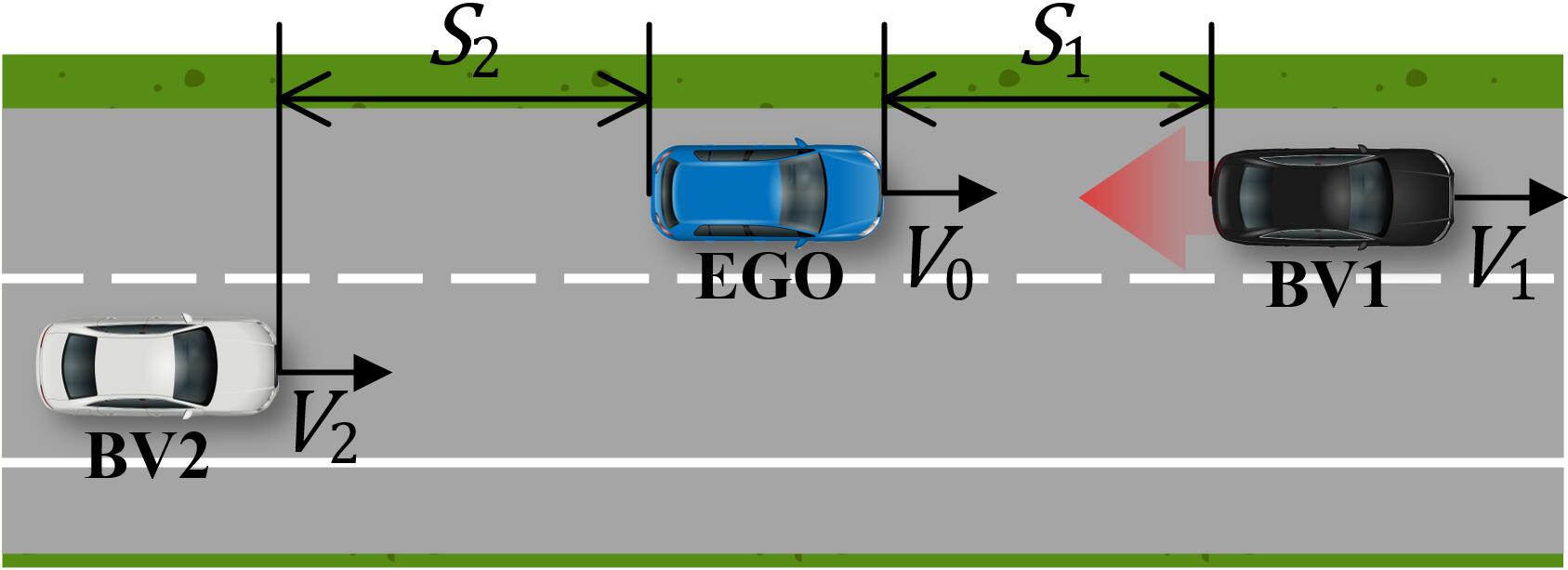}
      \caption{Test scenario.}
      \label{test_scenario}
\end{figure}

Car-following behavior is one of the basic driving behaviors and the core of traffic flow theories \cite{sun2023stability} and the car-following scenario is a typical test scenario for the safety evaluation of ADSs, especially when there exists a rear vehicle in the adjacent lane, in which the risk of the scenario stems from different vehicles, resulting in a multimodal risk distribution as discussed in Section \uppercase\expandafter{\romannumeral1}. The parameterized car-following scenario in this experiment is shown in Fig. \ref{test_scenario}. The ego vehicle (EGO) is driven by the built-in ADS of VTD (decision-making, planning, and control, with ideal perception) with an initial velocity $V_0$. While the background vehicle BV1 drives right in front of the ego vehicle with an initial distance $S_1$ and speed $V_1$, which keeps driving at a constant speed until 4 seconds of the simulation and then makes a full brake to stop. The other background vehicle BV2 drives behind the ego vehicle in the adjacent lane with distance $S_2$ and speed $V_2$. Throughout the whole scenario, the velocity of BV2 remains unchanged. Under the above conditions, the ego vehicle needs to brake or/and steer to avoid a collision with BV1, while making sure not to collide with BV2 when steering. Therefore, the distance $S_1$ between EGO and BV1 and the velocity $V_2$ of BV2 are two key parameters in this logical scenario, which are chosen as search parameters, while other scenario parameters take fixed default values. The range of the search parameters and the default value of the fixed parameters are shown in Table \ref{tab2}. To represent how close the ego vehicle is to collision, as well as to define an objective function for optimization, the criterion TTC \cite{vogel2003comparison} is used (Eq. (\ref{eq13})). Scenarios with minimum TTC during simulation smaller than 0.5s are considered critical, in which a crash is mostly unavoidable.

\begin{equation}\label{eq13}
TTC=\begin{cases}
    \frac{S_1}{V_0-V_1}, ~\text{if}\ S_1>0\ \text{and}\ V_0>V_1 \vspace{1ex}\\
    \infty, ~~~~~~~\text{if}\ V_0<V_1 \vspace{1ex} \\
    0, ~~~~~~~~~\text{if}\ S_1<0
\end{cases}
\end{equation}

\begin{table}[b]
\renewcommand{\arraystretch}{1.3}
\caption{Parameter Settings for Logical Scenario.\label{tab2}}
\centering
\begin{tabular}{ m{0.33\linewidth}  m{0.15\linewidth} m{0.2\linewidth} m{0.08\linewidth}}
\toprule
\textbf{Parameter} & \textbf{Denotation} & \textbf{Value/Range} & \textbf{Unit}\\
\midrule
Initial velocity of EGO & $V_0$  &  30 & $m/s$ \\

Initial velocity of BV1 & $V_1$  &  20 & $m/s$ \\

\textbf{Initial velocity of BV2} & $\boldsymbol{V_2}$  &  $ \boldsymbol{10 \sim 30} $  & $\boldsymbol{m/s}$ \\

\textbf{Initial distance between EGO and BV1} & $\boldsymbol{S_1}$  &  $ \boldsymbol{10 \sim 110} $ & $\boldsymbol{m}$\\

Initial distance between EGO and BV2 & $S_2$  &  50 & $m$\\

The moment when BV1 starts braking  & /  &  4 & $s$\\
\bottomrule
\end{tabular}
\end{table}

\begin{figure}[t] 
      \centering
      \includegraphics[width=.9\linewidth]{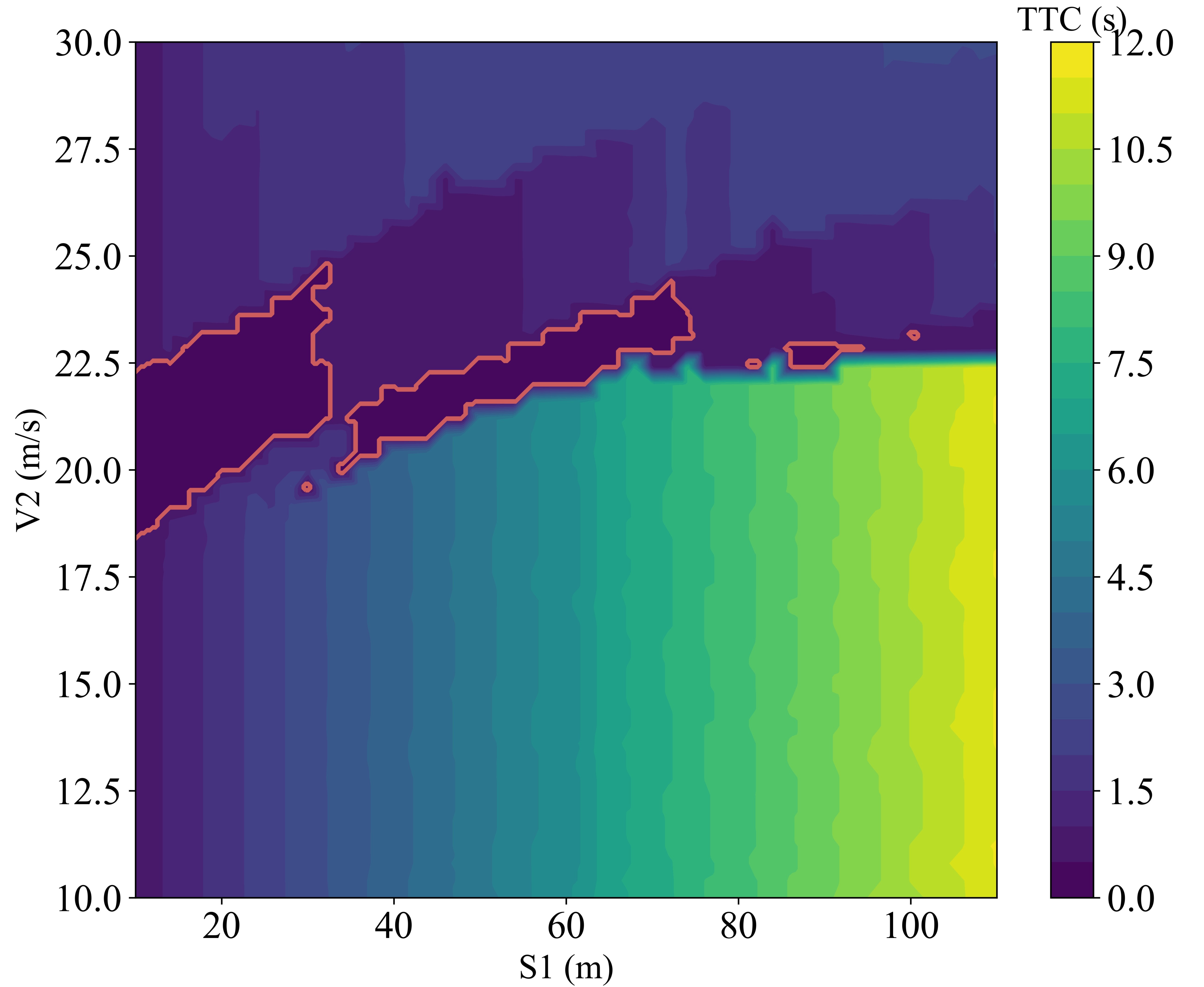}
      \caption{The ground truth of the logical scenario by exhaustive grid search.}
      \label{groundtruth_twoPara}
\end{figure}

To obtain the ground truth of the objective function in the logical scenario space, an exhaustive grid search is performed. The grid resolution is 100 $\times$ 100 and the ground truth is shown in Fig. \ref{groundtruth_twoPara}, with red color lining the critical subspaces (TTC $<$ 0.5s). The irregularly distributed critical subspaces suggest that a concrete scenario is critical only under certain parameter combinations. Thus, it is hard to judge whether a concrete scenario is critical only by knowledge or expert experience before testing. However, finishing such an exhaustive testing is time-consuming. In our experiment, it took 12 seconds to test one concrete scenario and totally cost around 33 hours to obtain Fig. \ref{groundtruth_twoPara}. The above information once again proves the necessity of using optimization algorithms to speedup the test. 

\subsubsection{Test Process and Results}

\begin{figure}[!b]%
    \centering
    \subfloat[F2 score of LAMBDA during 5000 samples]{
        \label{F2scoreTwopara}
        \includegraphics[width=\linewidth]{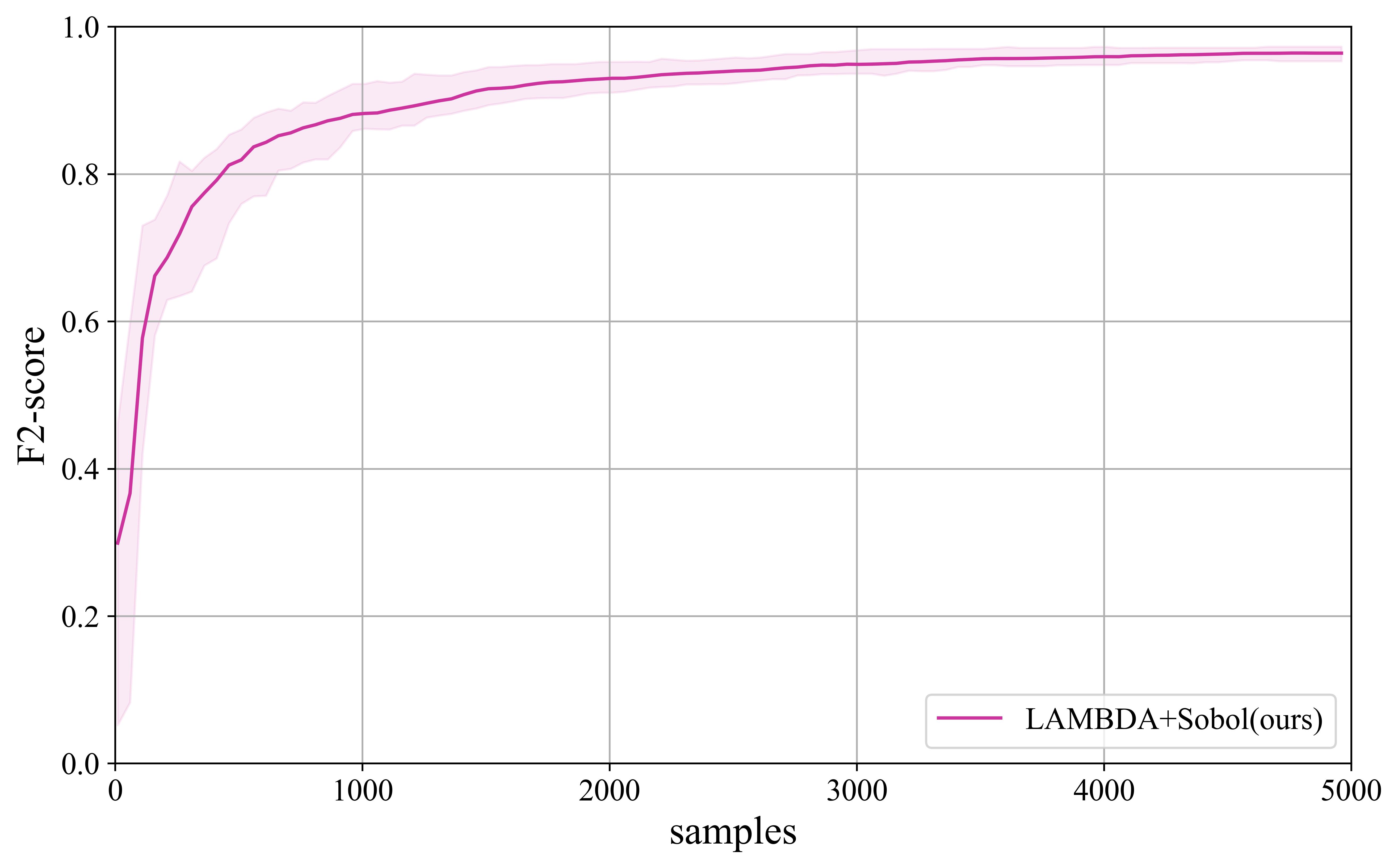}
        }\hfill
    \newline
    \subfloat[Sampling dynamics]{
        \label{samplingDynamics}
        \includegraphics[width=0.4\linewidth]{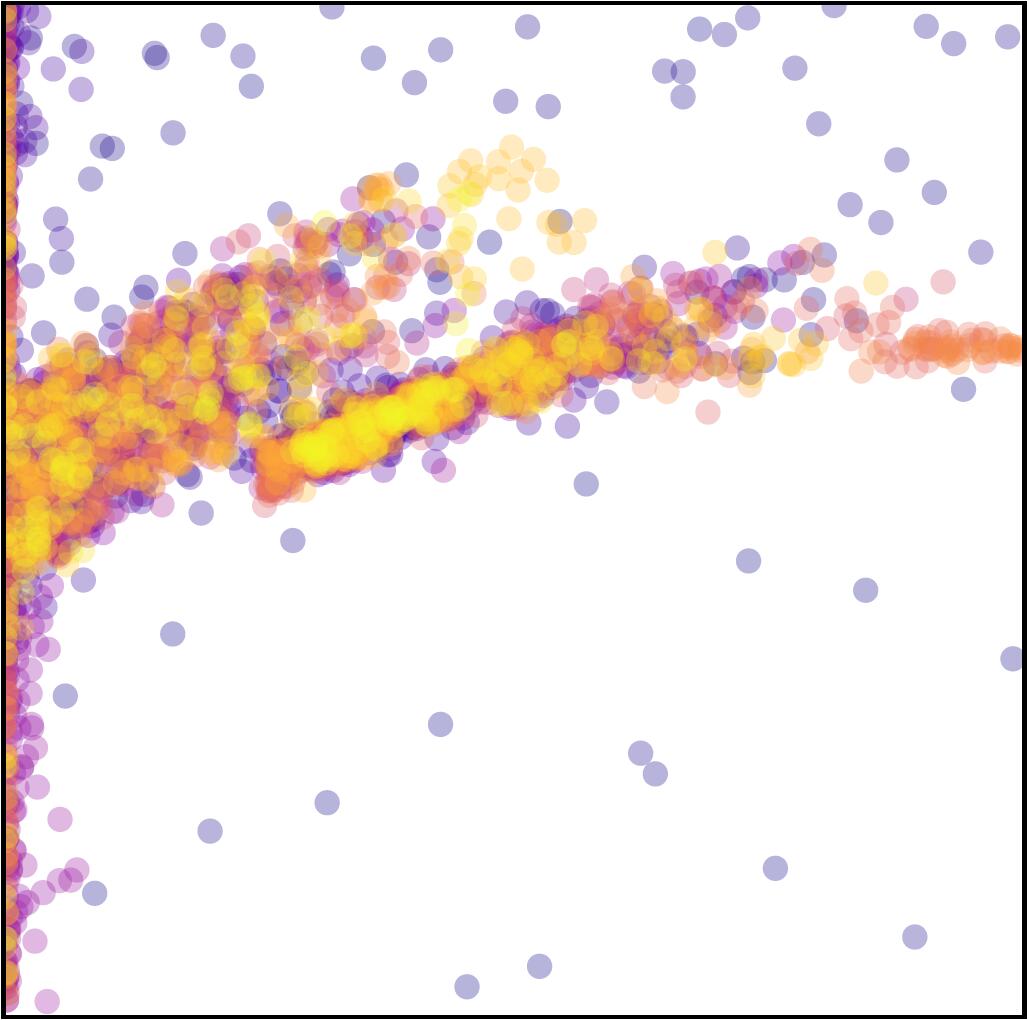}
        }\hfill
    \subfloat[Interpolation result]{
        \label{interpolation}
        \includegraphics[width=0.4\linewidth]{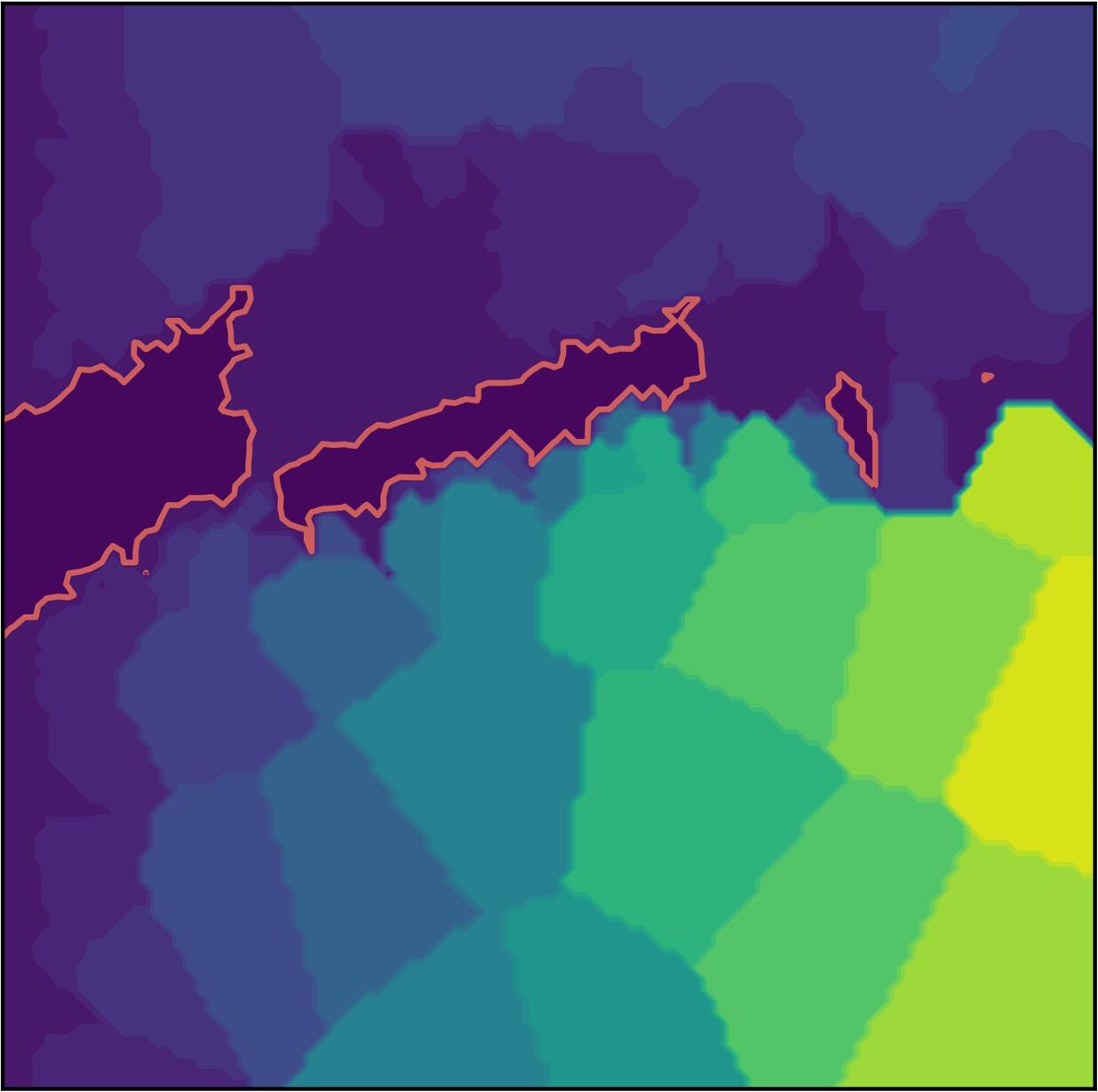}
        }
    \caption{Experimental results of the search of logical scenario. (a) F2 score of LAMBDA during 5000 samples. (b) Sampling dynamics of LAMBDA with 3000 samples. 
    (c) Interpolation result by regressor $\hat{f}$ based on the sampling records in Fig. \ref{samplingDynamics}. 
    }
    \label{res_twoPara}
\end{figure}

As depicted in Fig. \ref{LambdaProcess}, after the initial sampling by Sobol, LAMBDA decides where to sample next so as to find all critical scenarios as soon as possible. Once a sampling point has been determined, an executable OpenSCENARIO file is generated based on the combination of the logical scenario parameters and then run in the simulation software. During the simulation, the criterion TTC of the EGO is calculated simultaneously and the minimum TTC throughout the simulation, which can be regarded as the value of the objective function, will be sent back to LAMBDA. A smaller TTC means a higher risk level and therefore a higher test value. With all the results of the current batch of scenarios collected, UCB scores of each subspace are calculated and next batch of sampling points (namely the concrete scenarios) are then determined. The results of this experiment are shown in Fig. \ref{res_twoPara}. 

As can be seen from the F2 score in Fig. \ref{F2scoreTwopara}, LAMBDA covers all critical areas with high efficiency and stability, obtaining a final average value of 0.96. Furthermore, the sampling dynamics of LAMBDA with 3000 points is illustrated in Fig. \ref{samplingDynamics}. As the number of samples increases, new sampling points are apparently more concentrated in critical areas. Based on this, we train the regressor $\hat{f}$ mentioned in Section \ref{metric} and get the interpolation result shown in Fig. \ref{interpolation}. The three main critical areas interpolated by $\hat{f}$ have almost the same distribution as the ground truth in Fig. \ref{groundtruth_twoPara}, suggesting the accuracy of our algorithm, and the F2 score of 0.94 proves it as well. As for benchmarks with other baseline algorithms, detailed information can be found in Appendix \ref{Appendix_B}. It should be noted that although the logical scenario constructed in this experiment exhibits the multimodal nature of scenario risk, the almost connected critical areas and their large percentage of the entire logical scenario parameter space make it less challenging for the optimization algorithms. Therefore, even RS and Sobol can reach an F2 score above 0.9 within 5000 points in Appendix \ref{Appendix_B}. Since the selection of key parameters in this logical scenario is only determined by the knowledge of the authors while designing challenging logical scenarios is not the scope of this paper, the experiment in this section is only used to demonstrate the capabilities of LAMBDA for practical applications.

\section{Conclusion} 

In this paper, we propose the Black-Box Coverage problem for scenario-based virtual testing of ADSs, which means covering the solution set that satisfies an inequality containing a black-box function. To achieve as high coverage as possible within a limited optimization budget, we develop the LAMBDA algorithm based on the idea of search space quantization. LAMBDA utilizes density information to overcome sampling bias and introduces Beam Search to obtain more parallelizability. Moreover, to verify and evaluate the performance of our proposed algorithm under multimodal and high-dimensional situations, a metric F2 score to measure the coverage and a synthetic function Ripples to challenge the algorithm are proposed. The experimental results demonstrate the outstanding performance of LAMBDA on Black-Box Coverage problems. According to the benchmarks, LAMBDA can be 33 times on the 2-dimensional Holder-Table function and 6000 times on the 5-dimensional Ripples function faster than Random Search to get 95\% coverage. Furthermore, the experiment in practical application shows that LAMBDA has great potential in the safety evaluation of ADSs. In future work, we plan to adopt LAMBDA with high-dimensional and more challenging logical scenarios \cite{yang2023adaptive}. Moreover, instead of TTC, more comprehensive scenario risk indicators, such as DNDA \cite{wu2022risk} will be used as the objective function to obtain a more reasonable critical scenario set.

\appendices
\section{Hyper-Parameter Settings of LAMBDA and Baseline Algorithms} \label{Appendix_A}
The hyper-parameters in baselines are set according to the following criteria:
\begin{itemize}
\item{If the author of an algorithm has given a suggestion about how to set the hyper-parameters in the paper or in the code, we adopt the suggested setting.}
\item{If no suggestion is given about a hyper-parameter, but there is a similar hyper-parameter in LAMBDA, we keep the baseline’s setting the same as that of LAMBDA.}
\item{Otherwise, the default hyper-parameters in the baseline’s codes are used.}
\end{itemize}

Detailed hyper-parameter settings of LAMBDA and all baselines in Holder-Table and Ripples function benchmarks are shown in Table \ref{AppendTable1} and Table \ref{AppendTable2}.

\section{Detailed Information about the Benchmarks on the Practical Issue} \label{Appendix_B}

\begin{figure}[h] 
      \centering
      \includegraphics[width=\linewidth]{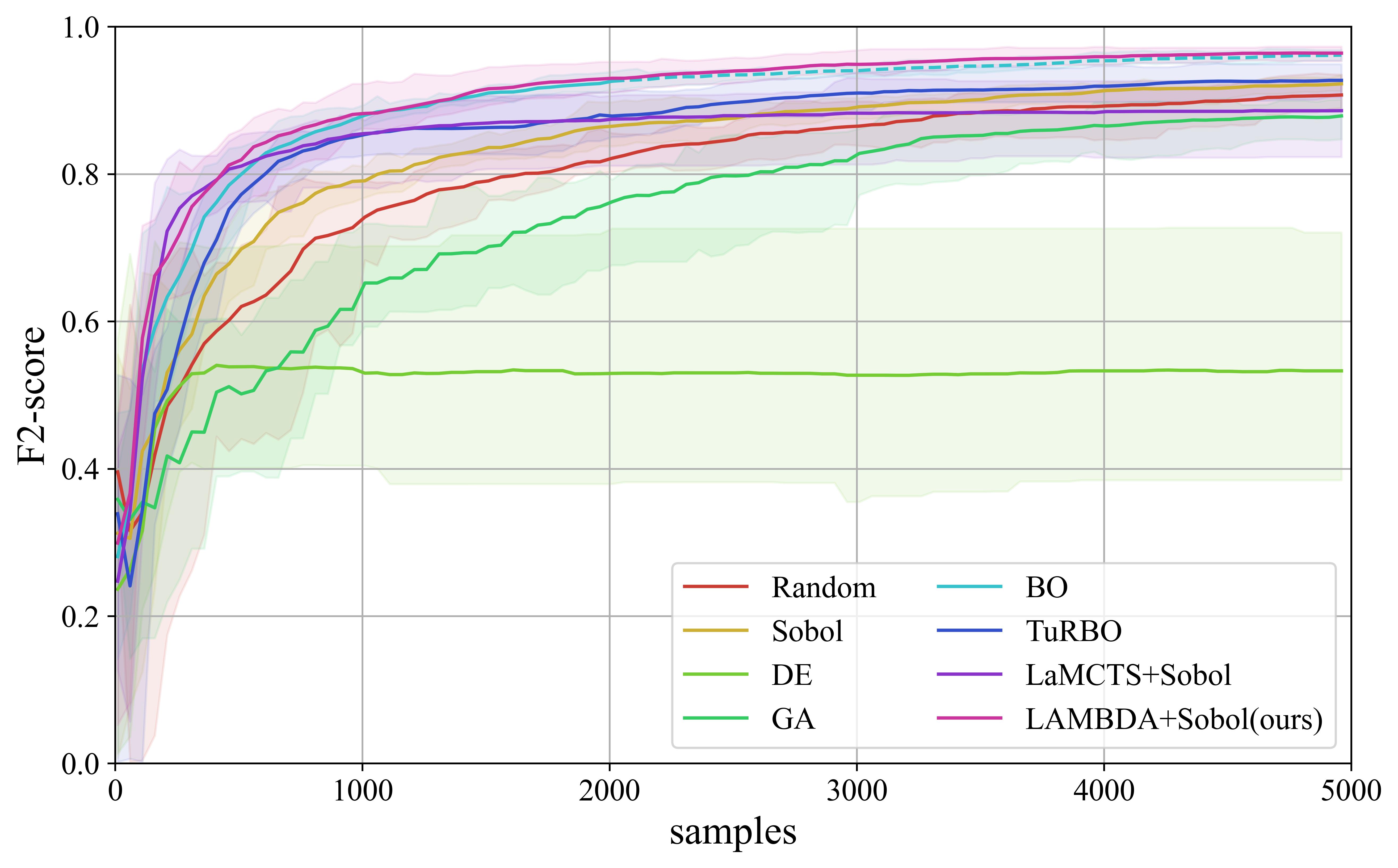}
      \caption{Benchmarks on practical issue.}
      \label{benchmark_practical}
\end{figure}

The calculated F2 scores of all algorithms are shown in Fig. \ref{benchmark_practical}. Detailed hyper-parameter settings of LAMBDA and all baselines in the benchmarks on the practical issue are shown in Table \ref{AppendTable3}. For simplicity of description, only the parameters that differ from those in Table \ref{AppendTable1} are listed, the rest of the parameters remain unchanged from Table \ref{AppendTable1}. 

\begin{table*}[!h]
\renewcommand{\arraystretch}{1.3}
\caption{Hyper-Parameter Settings for Holder-Table function benchmarks.\label{AppendTable1}}
\centering
\begin{threeparttable}
\begin{tabular}{m{0.1\linewidth}  m{0.25\linewidth} m{0.2\linewidth} m{0.3\linewidth}}
\toprule
\textbf{Algorithm}  & \textbf{Hyper-parameter} & \textbf{Setting} & \textbf{Description} \\
\midrule
\multirow{8}{*}{LAMBDA} & exploration factor $c_p$ & 1 & /  \\ \cline{2-4} 
~ & stop criterion $leafsize$ & 10 & / \\ \cline{2-4}
~ & stop criterion $depth$ & 8 & / \\ \cline{2-4}
~ & initial sampling points & 256 & by Sobol \\ \cline{2-4}
~ & optimization budget $N$ & 5000 & / \\ \cline{2-4}
~ & beam width & 2 & larger beams with smaller $c_p$ \\ \cline{2-4}
~ & selections per treeification & 50 & /  \\ \cline{2-4}
~ & samples per selection & 1 & Sobol as the local sampler \\ \hline

\multirow{7}{*}{LaMCTS} & exploration factor $c_p$ & 0.1${y_{max}}^*$ \newline e.g. $0.1 \times 20 = 2 $& suggestion from \cite{wang2020learning} \\ \cline{2-4} 
~ & stop criterion $leafsize$ & 10 & \multirow{6}{*}{the same as LAMBDA} \\ \cline{2-3}
~ & stop criterion $depth$ & 8 & ~ \\ \cline{2-3}
~ & initial sampling points & 256 & ~ \\ \cline{2-3}
~ & optimization budget $N$ & 5000 & ~ \\ \cline{2-3}
~ & selections per treeification & 50 & ~ \\ \cline{2-3}
~ & samples per selection & 1 & ~ \\ \hline

\multirow{3}{*}{BO} & initial sampling points & 256  & \multirow{3}{*}{the same as LAMBDA} \\ \cline{2-3}
~ & optimization budget $N$ & 5000 & ~ \\ \cline{2-3}
~ & samples per Thompson Sampling & 50 & ~ \\ \hline

\multirow{4}{*}{TuRBO-4} & initial sampling points & 256 \newline (64 per Trust Region) & \multirow{3}{*}{the same as LAMBDA} \\ \cline{2-3}
~ & optimization budget $N$ & 5000 & ~ \\ \cline{2-3}
~ &  samples per Thompson Sampling & 50 & ~ \\ \cline{2-4}
~ & number of Trust Regions & 4 & according to the experimental results from \cite{eriksson2019scalable} \\ \hline

\multirow{3}{*}{GA} & population size & 50  & \multirow{2}{*}{default setting in the lib \textit{scikit-opt}} \\ \cline{2-3}
~ & mutation probability & 0.001 & ~ \\ \cline{2-4}
~ & optimization budget $N$ & 5000 & the same as LAMBDA \\ \hline

\multirow{3}{*}{DE} & population size & 50  & \multirow{2}{*}{default setting in the lib \textit{scikit-opt}} \\ \cline{2-3}
~ & mutation probability & 0.3 & ~ \\ \cline{2-4}
~ & optimization budget $N$ & 5000 & the same as LAMBDA \\ \hline

Sobol & / & / &/ \\ \hline
Random Search & / & / & / \\

\bottomrule
\end{tabular}
\begin{tablenotes}    
    \footnotesize               
    \item * $y_{max}$ is the maximum value of the objective function.
\end{tablenotes}
\end{threeparttable}
\end{table*}

\begin{table*}[!h]
\renewcommand{\arraystretch}{1.3}
\caption{Hyper-parameter Settings for 5-dimensional Ripples Function Benchmarks.\label{AppendTable2}}
\centering
\begin{tabular}{m{0.1\linewidth}  m{0.25\linewidth} m{0.2\linewidth} m{0.3\linewidth}}
\toprule
\textbf{Algorithm}  & \textbf{Hyper-parameter} & \textbf{Setting} & \textbf{Description} \\
\midrule
\multirow{17}{1.5cm}{LAMBDA with BO/TuRBO} & exploration factor $c_p$ & 0.8 & /  \\ \cline{2-4} 
~ & stop criterion $leafsize$ & 50 & / \\ \cline{2-4}
~ & stop criterion $depth$ & 9 & / \\ \cline{2-4}
~ & initial sampling points & 1024 & by Sobol \\ \cline{2-4}
~ & optimization budget $N$ & 50,000 & / \\ \cline{2-4}
~ & beam width & 15 & larger beams with smaller $c_p$ \\ \cline{2-4}
~ & selections per treeification & 90 & / \\ \cline{2-4}
~ & samples per selection & / & depends on the local sampler \\ \cline{2-4}
~ & \multicolumn{3}{l}{\textbf{Hyper-parameter settings for the local sampler: BO} } \\  \cline{2-4}
~ & initial sampling points & / & depends on the number of historical sampling points in the current subspace \\ \cline{2-4}
~ & local optimization budget $N$ & 50 & /  \\ \cline{2-4}
~ & samples per Thompson Sampling & 50 & / \\ \cline{2-4}
~ & \multicolumn{3}{l}{\textbf{Hyper-parameter settings for the local sampler: TuRBO-1} } \\  \cline{2-4}
~ & initial sampling points & 30 & by Latin hypercube sampling in the Trust Region with Reject sampling \\ 
\bottomrule
\end{tabular}

\end{table*}

\begin{table*}[!h]
\renewcommand{\arraystretch}{1.3}
\centering
\begin{threeparttable}
\begin{tabular}{m{0.1\linewidth}  m{0.25\linewidth} m{0.2\linewidth} m{0.3\linewidth}}
\toprule
\textbf{Algorithm}  & \textbf{Hyper-parameter} & \textbf{Setting} & \textbf{Description} \\
\midrule
~ & local optimization budget $N$ & / & local sampling ends when the length of the TR is less than a predefined threshold  \\ \cline{2-4}
~ & samples per Thompson Sampling & 5 & / \\ \cline{2-4}
~ & number of Trust Regions & 1 & the same as the settings in \cite{wang2020learning} \\ \hline

~ & exploration factor $c_p$ & $0.8^*$  & \multirow{6}{*}{the same as LAMBDA} \\ \cline{2-3} 
~ & stop criterion $leafsize$ & 50 & ~ \\ \cline{2-3}
\multirow{5}{1.5cm}{LaMCTS with BO/TuRBO} & stop criterion $depth$ & 9 & ~ \\ \cline{2-3}
~ & initial sampling points & 1024 & ~ \\ \cline{2-3}
~ & optimization budget $N$ & 50000 & ~ \\ \cline{2-3}
~ & selections per treeification & 90 & ~ \\ \cline{2-4}
~ & samples per selection & / & depends on the local sampler \\ \cline{2-4}
~ & \multicolumn{2}{l}{\textbf{Hyper-parameter settings for the local sampler: BO} } &  \multirow{2}{*}{the same as LAMBDA} \\  \cline{2-3}
~ & \multicolumn{2}{l}{\textbf{Hyper-parameter settings for the local sampler: TuRBO-1} } & ~ \\  \hline

\multirow{5}{*}{TuRBO-5} & initial sampling points & 1025 \newline (205 per Trust Region) & \multirow{2}{*}{the same as LAMBDA} \\ \cline{2-3}
~ & optimization budget $N$ & 50000 & ~ \\ \cline{2-4}
~ & samples per Thompson Sampling & 50 & / \\ \cline{2-4}
~ & number of Trust Regions & 5 & / \\

\bottomrule
\end{tabular}
\begin{tablenotes}    
    \footnotesize               
    \item * We find the suggestion of $c_p$ from \cite{wang2020learning} isn't suitable for this case.
\end{tablenotes}
\end{threeparttable}
\end{table*}

\begin{table*}[!h]
\renewcommand{\arraystretch}{1.3}
\caption{Hyper-Parameter Settings for the Practical Issue.\label{AppendTable3}}
\centering
\begin{tabular}{m{0.1\linewidth}  m{0.25\linewidth} m{0.2\linewidth} m{0.3\linewidth}}
\toprule
\textbf{Algorithm}  & \textbf{Hyper-parameter} & \textbf{Setting} & \textbf{Description} \\
\midrule
\multirow{2}{*}{LAMBDA} & exploration factor $c_p$ & 0.25 & /  \\ \cline{2-4} 
~ & initial sampling points & 64 & by Sobol \\ \hline

\multirow{2}{*}{LaMCTS} & exploration factor $c_p$ & 1 & suggestion from \cite{wang2020learning} \\ \cline{2-4} 
~ &  initial sampling points & 64 & the same as LAMBDA \\ \hline

BO & initial sampling points & 64 & the same as LAMBDA \\ \hline

TuRBO-4 & initial sampling points & 64 \newline (16 per Trust Region) & the same as LAMBDA \\ \hline

GA & \multicolumn{3}{l}{\multirow{4}{*}{the same as the settings in Table \ref{AppendTable1}}}  \\ \cline{1-1}
DE & ~ & ~  & ~ \\ \cline{1-1}
Sobol & ~ & ~ &~ \\ \cline{1-1}
Random Search & ~ & ~ & ~ \\
\bottomrule
\end{tabular}

\end{table*}

\bibliographystyle{IEEEtran}
\bibliography{citelist}

\vfill

\end{document}